\documentclass[letterpaper]{article} % DO NOT CHANGE THIS
\usepackage{aaai2026}
\usepackage{times}  % DO NOT CHANGE THIS
\usepackage{helvet}  % DO NOT CHANGE THIS
\usepackage{courier}  % DO NOT CHANGE THIS
\usepackage[hyphens]{url}  % DO NOT CHANGE THIS
\usepackage{graphicx} % DO NOT CHANGE THIS
\DeclareGraphicsExtensions{.pdf,.png,.jpg}
\usepackage{natbib}  % DO NOT CHANGE THIS AND DO NOT ADD ANY OPTIONS TO IT
\usepackage{caption} % DO NOT CHANGE THIS AND DO NOT ADD ANY OPTIONS TO IT
\usepackage{algorithm}
\usepackage{algorithmic}

\usepackage{newfloat}
\usepackage{listings}

\usepackage{paralist}
\usepackage{amsmath}
\usepackage{amssymb}
\usepackage{multirow}
\usepackage{threeparttable}
\usepackage{subcaption}
\usepackage{booktabs}

\usepackage[many]{tcolorbox}
\usepackage[dvipsnames, svgnames, x11names]{xcolor}
\usepackage{tikz}
\newcommand{\encircle}[1]{\tikz[baseline=(char.base)]{
    \node[shape=circle,draw,inner sep=1pt] (char) {#1};}}

\tcbuselibrary{listingsutf8}
\newtcolorbox[auto counter, number within=section]{case}[2][]{colframe=green!70!black, colback=yellow!20!white, coltitle=black, fonttitle=\bfseries, title=Case~\thetcbcounter: #2,#1}
\newtcolorbox[auto counter, number within=section]{prompt}[2][]{colframe=blue!75!black, colback=yellow!20!white, fonttitle=\bfseries}

\DeclareCaptionStyle{ruled}{labelfont=normalfont,labelsep=colon,strut=off} % DO NOT CHANGE THIS
\floatstyle{ruled}
\newfloat{listing}{tb}{lst}{}
\floatname{listing}{Listing}
\nocopyright
\title{D-SMART: Enhancing LLM Dialogue Consistency via \\Dynamic Structured Memory And Reasoning Tree}
\author {
    Xiang Lei \textsuperscript{\rm 1},
    Qin Li\textsuperscript{\rm 1}\thanks{Correspondence to: Qin Li \text{$<$}qli@sei.ecnu.edu.cn\text{$>$}.},
    Min Zhang\textsuperscript{\rm 1},
    Min Zhang\textsuperscript{\rm 1}
}

\affiliations {
    \textsuperscript{\rm 1} Shanghai Key Laboratory of Trustworthy Computing, \\
    East China Normal University,\\
    Shanghai, China
}

\usepackage{bibentry}
\begin{document}

\maketitle

\begin{abstract}
Large Language Models (LLMs) often exhibit factual inconsistencies and logical decay in extended, multi-turn dialogues, a challenge stemming from their reliance on static, pre-trained knowledge and an inability to reason adaptively over the dialogue history. Prevailing mitigation strategies, such as Retrieval-Augmented Generation (RAG) and agentic working memories, improve information recall but still engage with fundamentally static knowledge sources and follow pre-defined single reasoning path. This hinders their ability to preserve factual and logical consistency of their responses in multi-turn dialogues while the context evolves over time. To address this issue, we propose D-SMART, a model-agnostic framework designed to maintain multi-turn dialogue consistency by enabling LLMs to build and reason over a dynamic, structured representation of the conversational context. This is achieved via two synergistic components: (1) a Dynamic Structured Memory (DSM), which incrementally constructs and maintains an authoritative, OWL-compliant knowledge graph of the conversation; and (2) a Reasoning Tree (RT), which executes inferences as an explicit and traceable multi-step search over the graph. As the popular-used quality score (judged by GPT-4) can overlook logical flaws, we introduce new NLI-based metrics to better measure multi-turn dialogue consistency. Comprehensive experiments on the MT-Bench-101 benchmark show that D-SMART significantly outperforms state-of-the-art baselines, elevating the dialogue consistency score by over 48\% for both proprietary and open-source models, and notably improves the quality score of the latter by up to 10.1\%. 
\end{abstract}

\section{Introduction}
While Large Language Models (LLMs) have demonstrated remarkable capabilities across various tasks \cite{li2024fundamental}, maintaining logical and factual consistency in extended, multi-turn dialogues remains a significant challenge \cite{yi2024survey}. Lapses in this consistency can severely undermine effective information conveyance, intention clarification, and consensus building \cite{she-etal-2024-exploring}. However, the architectural foundations of contemporary LLMs often struggle to meet this critical requirement, especially as dialogues become longer and more complex \cite{liu2023exposing, 10.1145/3703155}.

\begin{figure}[ht]
\centering
\includegraphics[width=0.98\columnwidth]{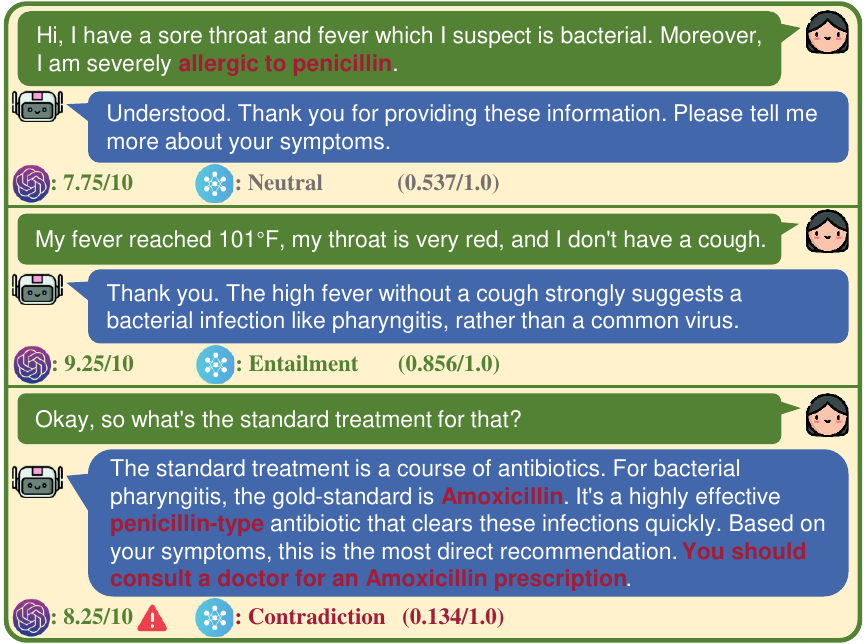}
\caption{A case of LLM exhibiting logical inconsistency in a multi-turn dialogue. }
\label{fig:case}
\end{figure}

This challenge stems from a fundamental reliance on static, pre-trained knowledge and the processing of an unstructured dialogue history, where attention to prior turns can decay \cite{hu2024towards, zhao-etal-2024-enhancing}. This can lead to difficulties in resolving context ambiguity, managing conflicts between the model's internal knowledge and newly introduced dialogue-specific facts \cite{xu-etal-2024-knowledge-conflicts}, and preventing the propagation of errors from early misunderstandings \cite{laban2025llms}.

Such vulnerabilities can lead to critical failures, as illustrated in Figure~\ref{fig:case}, where the model contradicts prior user input. This exposes the limitation of prevailing mitigation strategies like RAG \citep{gao2023retrieval} or agentic working memories. While such methods augment LLMs with external knowledge \cite{she-etal-2024-exploring} or textual history, they do not guarantee that the model will correctly reason over the emergent facts presented \cite{xue2023rcot}. This gap hinders their ability to maintain long-term logical coherence and adapt to evolving factual contexts \cite{liu-etal-2024-lost}. Notably, an eloquent but inconsistent response, as shown in Figure~\ref{fig:case}, may still receive a high score from holistic quality evaluators like GPT-4, which can overlook severe logical flaws. This highlights the pressing need for more discerning evaluation methods that can directly assess the logical integrity of dialogue responses.

To address these limitations in both reasoning and evaluation, we propose \textbf{D-SMART}, a framework comprised of two synergistic components: (1) a \textbf{Dynamic Structured Memory (DSM)}, which incrementally constructs an OWL-compliant knowledge graph to serve as a dynamic, logically-consistent knowledge base tailored for the dialogue; and (2) a \textbf{Reasoning Tree (RT)}, which guides the LLM to perform explicit, traceable, multi-step reasoning over the DSM. In essence, the DSM provides a structured memory of ``what to think about", while the RT provides a deliberate process for ``how to think" based on that memory. Complementing the framework, we also introduce new Natural Language Inference (NLI)-based metrics to better assess multi-turn dialogue consistency, moving beyond holistic quality scores. 

The main contributions of this paper are as follows:
\begin{itemize}
    \item We propose \textbf{D-SMART}, a model-agnostic framework that drives traceable and faithful reasoning over a dynamic, dialogue-specific structured memory, enhancing dialogue consistency.

    \item We introduce two NLI-based metrics, the \textbf{Consistency Score (CS)} and the \textbf{Dialogue Entailment Rate (DER)}, providing a more precise evaluation of multi-turn dialogue consistency by measuring logical entailment.
    
    \item Comprehensive experiments on MT-Bench-101 show D-SMART significantly outperforms baselines, boosting consistency by over 48\% and demonstrating stability against logical decay in extended dialogues turns.
\end{itemize}

\begin{figure}[ht]
\centering
\includegraphics[width=0.98\columnwidth]{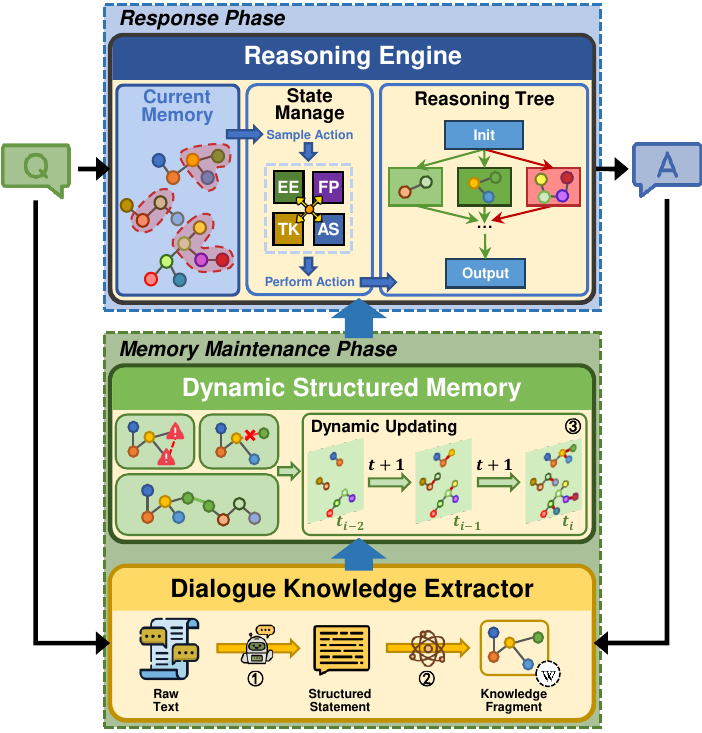}
\caption{An overview of the D-SMART framework. }
\label{fig:overview}
\end{figure}

\section{The D-SMART Framework}
Inspired by cognitive architectures like Soar \cite{LAIRD19871}, we propose D-SMART, a framework to enhance the logical consistency and factual grounding of LLMs in multi-turn dialogues. Formally, instead of the traditional paradigm: $r_t \sim p_\theta(r_t \mid q_t, \mathcal{H}_{t-1})$, D-SMART reformulates response generation as
\begin{align}
\label{eq:final_answer_from_D_SMART}
r_t &\sim p_\theta(r_t\mid q_t,\mathcal{H}_{t-1}, \mathcal{G}_{t-1}, \mathcal{T}_t^*)
\end{align}
thereby coupling a dialogue-specific memory graph $\mathcal{G}_{t}$ with a reasoning tree $\mathcal{T}_t^*$. As illustrated in Figure \ref{fig:overview}, in the \textbf{Response Phase}, upon receiving a user query ($q_t$), the framework forwards it to the Reasoning Engine. Guided by a set of actions, the engine's RT actively explores the DSM ($\mathcal{G}_{t-1}$) to construct a grounded context for response generation ($r_{t}$). Subsequently, the completed turn ($q_t, r_t$) triggers the \textbf{Memory Maintenance Phase}, where the framework extracts new knowledge and resolves any emerging conflicts to dynamically update the memory graph. This updated DSM ($\mathcal{G}_{t}$) is prepared for the subsequent Response Phase, ensuring that the model's responses are continuously anchored in an evolving representation of the conversational common ground.

\subsection{Dynamic Structured Memory Construction and Maintenance}
The DSM mitigates context decay by incrementally transforming the unstructured dialogue history into an evolving, OWL-compliant knowledge graph. Unlike standard dialogue systems that represent history as a flat text sequence $\mathcal{H}_{t-1}$ \cite{logan-etal-2019-baracks}, which can be susceptible to information loss and contradiction \cite{liu-etal-2022-relational}, the DSM provides a persistent and computationally tractable knowledge base. 

Formally, this graph at turn $t$, which is a set of factual triples $\mathcal{G}_{t} = \{(e_{h}, r, e_{t})\} \subseteq \mathcal{E}_{t} \times \mathcal{R}_{t} \times \mathcal{E}_{t}$, yields a structured representation of the conversational common ground \cite{cheng2023lift}, built exclusively from the dialogue itself.

The construction and maintenance of the DSM follows a turn-by-turn, multi-step workflow designed to ensure semantic fidelity and logical coherence \cite{hixon-etal-2015-learning}. For each completed dialogue turn $h_t = (q_t, r_t)$, the following steps are executed:

\begin{enumerate}
    \item \textbf{Structured Statement Generation:} We first employ an LLM to distill the core semantic content of the turn $h_t$ into a concise, assertion-like natural language statement \cite{luo2023reasoning}, $s_t$, normalizing colloquial or fragmented utterances into a format amenable to knowledge extraction, which can be formalized as: $s_t \sim p_\theta^{s}(s_t \mid h_t)$, where $p_\theta^s$ is the LLM prompted to perform this distillation (prompt details are provided in Appendix E).

    \item \textbf{Knowledge Fragment Extraction:} From $s_t$, we extract a OWL-compliant knowledge graph fragment, $\mathcal{G}_t^\prime$. This is achieved by a neuro-symbolic pipeline \cite{10.1007/978-3-031-94578-6_4}, denoted as $\mathrm{KGE}$: $\mathcal{G}_t^\prime = \mathrm{KGE}(s_t)$, achieving approximately 95\% overall accuracy (Algorithm 1 in Appendix A). OWL’s formal semantics provide a foundation for consistency checking and logical inference, thereby safeguarding against the propagation of logical errors \citep{DEGIORGIS2025104127}. 
    % The specific implementation of KGE pipeline is detailed in Appendix A.
    
    \item \textbf{Dynamic Update and Conflict Resolution:} To account for new, clarified, or corrected information, the newly extracted fragment $\mathcal{G}_t^\prime$ is merged with the existing graph $\mathcal{G}_{t-1}$. This is a two-stage conflict resolution process:
    \begin{itemize}
        \item \textbf{Conflict Detection:} An LLM is prompted to perform a semantic comparison between the serializations of $\mathcal{G}_{t-1}$ and $\mathcal{G}_t^\prime$. It identifies any triples in $\mathcal{G}_{t-1}$ that are contradicted or superseded by the information in $\mathcal{G}_t^\prime$.
        \item \textbf{Graph Pruning and Merging:} Conflicting triples identified in the previous step are pruned from $\mathcal{G}_{t-1}$. The $\mathcal{G}_t^\prime$ is then merged into $\mathcal{G}_{t-1}$. An OWL reasoner can then be used to verify the logical consistency of the resulting graph $\mathcal{G}_t$ against predefined axioms.
    \end{itemize}
\end{enumerate}

This update mechanism, denoted by the operator $\oplus$, can be formalized as:
\begin{align}
\label{eq:update_dsm}
\mathcal{G}_t &= \mathcal{G}_{t-1} \oplus \mathcal{G}_t^\prime = (\mathcal{G}_{t-1} \setminus p_\theta^{c}(\mathcal{G}_{t-1}, \mathcal{G}_t^\prime)) \cup \mathcal{G}_t^\prime
\end{align}
where $p_\theta^c$ represents the LLM-based conflict detection function that returns the set of triples from $\mathcal{G}_{t-1}$ to be removed (see Appendix E for prompt details).

Through this workflow, the DSM evolves as a faithful and logically consistent chronicle of the dialogue. This structured memory provides a stable foundation for the system's reasoning capabilities. 

\begin{figure*}[htb]
\centering
\includegraphics[width=\textwidth]{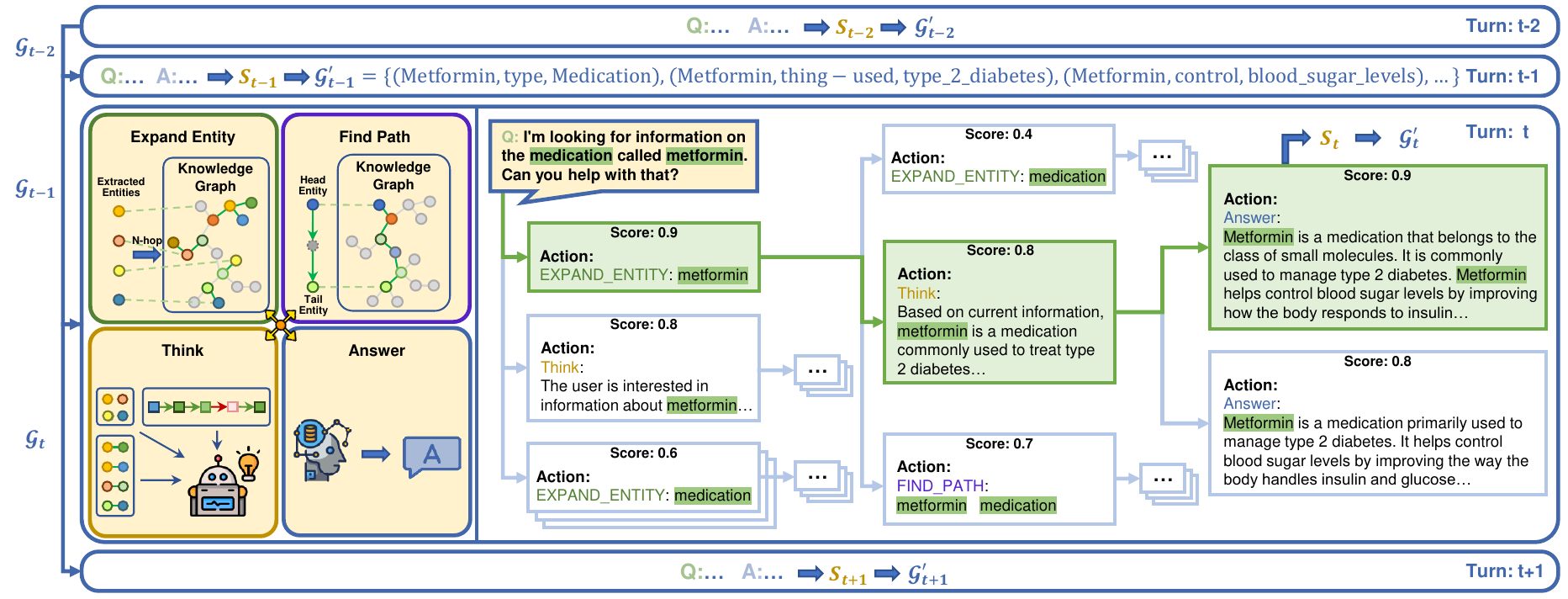}
\caption{The reasoning process at turn $t$. The tree expands by executing actions, enriching its local knowledge from the DSM. The LLM evaluates each step, and the final path leads to a grounded answer, which will update the DSM to $\mathcal{G}_t$ for the next turn.}
\label{fig:reasoning_process}
\end{figure*}

\subsection{Reasoning Tree for Memory Traversal}
Simply linearizing a graph for the LLM's context can be suboptimal, as LLMs may struggle to interpret raw structured data, which can lead to opaque and less reliable reasoning \cite{agrawal-etal-2024-knowledge,jiang-etal-2023-structgpt}. Effectively leveraging this structure for response generation requires a dedicated mechanism \cite{modarressi2023ret}, the RT guides the LLM to explore multiple, alternative reasoning paths over the DSM. Inspired by the Tree-of-Thought (ToT) paradigm \cite{yao2023tree}, the RT framework moves beyond a single, linear reasoning chain that may fail. This structure allows the model to deliberately explore diverse reasoning trajectories, backtrack from unpromising avenues, and select the most coherent path grounded in the DSM.

The core idea is to model the reasoning process for a given query $q_t$ as a search for an optimal path within a reasoning tree, $\mathcal{T}_t$. The construction of this tree is an iterative process where the LLM acts as a high-level planner. Starting from a root node representing the initial query, the framework progressively expands the tree by generating potential next steps. A search algorithm navigates this expanding tree, prioritizing more promising paths based on a value function, also estimated by the LLM. This mitigates the risk of error accumulation often seen in linear reasoning chains.

Formally, each node $\tau_i$ in the tree $\mathcal{T}_t$ represents a reasoning state $\mathcal{S}_i$. We define this state as a four-tuple: $\mathcal{S}_i = (\tilde{\mathcal{G}}_i, \mathcal{Z}_i, v_i, d_i)$.
\begin{itemize}
    \item $\tilde{\mathcal{G}}_i \subseteq \mathcal{G}_t$ is the relevant subgraph accumulated at node $\tau_i$. It serves as the grounded, local context for the current reasoning step.
    \item $\mathcal{Z}_i = \langle z_1, z_2, \ldots, z_j \rangle$ is the reasoning trajectory, a sequence of actions that led to the current state.
    \item $v_i \in [0, 1]$ is a value score assigned by the LLM, evaluating the potential of state $\mathcal{S}_i$ to lead to a correct answer for $q_t$. This is estimated via $v_i \sim p_\theta^{v}(v \mid q_t, \mathcal{S}_i)$.
    \item $d_i$ is the depth of the node in the tree.
\end{itemize}

To ground this reasoning process in the DSM, we define a set of discrete actions $\mathcal{A}$ that the LLM can propose to transition between states. These actions are designed for explicit graph traversal and knowledge manipulation, extending standard thought generation with symbolic operations:
\begin{itemize}
    \item \textbf{Expand Entity ($a_{ee}$)}: Given an entity in $\tilde{\mathcal{G}}_i$, retrieve its n-hop neighborhood from the full DSM $\mathcal{G}_t$ to enrich the local context.
    \item \textbf{Find Path ($a_{fp}$)}: Given two entities, search for connecting paths within $\mathcal{G}_t$ and add them to $\tilde{\mathcal{G}}_i$, facilitating multi-hop reasoning.
    \item \textbf{Think ($a_{tk}$)}: Synthesize the information in $\mathcal{S}_i$ to generate an intermediate thought or plan the subsequent action.
    \item \textbf{Answer ($a_{as}$)}: A terminal action to generate the final response based on the information gathered in $\mathcal{S}_i$.
\end{itemize}

At each expansion step from a node $\tau_i$, the LLM-based policy proposes an action $a_{ij} \sim p_\theta^\pi(a \mid q_t, \mathcal{S}_i)$, which results in a new state $\mathcal{S}_j$. The objective is to construct a tree that maximizes the joint probability of the sequence of actions forming the most plausible reasoning path. This can be formalized as:
\begin{align}
\label{eq:prob_reasoning_tree}
\mathcal{T}_t^* &= \arg\max_{\mathcal{T}_t} p_\theta(\mathcal{T}_t \mid q_t, \mathcal{G}_{t}) \\
&\nonumber\propto \prod\limits_{(\tau_i, a_{ij}, \tau_j) \in \mathcal{T}_t} p_\theta^\pi(a_{ij}\mid q_t, \mathcal{S}_i)
\end{align}

where the probability of a tree is defined by the product of policy probabilities for the sequence of actions taken.

The search is not a blind exploration but a traceable traversal over a structured memory, enabling a form of deliberate, ``slow" reasoning \cite{kahneman_thinking_2012} to address complex queries while ensuring conclusions remain faithful to the established dialogue history.

\subsection{From Structured Memory to Grounded Response}
The core design principle of D-SMART is to leverage the LLM not as a passive generator operating on unstructured text, but as a high-level semantic orchestrator. The LLM guides the structured traversal of the DSM via the RT framework to find an optimal reasoning path from the tree $\mathcal{T}_t^*$ that produces a coherent and grounded answer to the query $q_t$, transforming the opaque, internal reasoning of a standard LLM into an explicit and traceable process. 

To solve this search problem tractably, we operationalize the tree traversal using a beam search algorithm, which navigates the state space defined by the RT, balancing traversal of diverse reasoning paths with computational feasibility. At each step, the LLM-as-policy proposes actions from $\mathcal{A}$ to expand the most promising nodes on the beam. Graph-related actions are executed deterministically against the DSM, ensuring that each reasoning step is grounded in the structured memory. The LLM-as-value-function then evaluates the newly generated states, and the beam is updated to retain the top-$k$ most promising candidates for the next iteration. The search terminates upon reaching a maximum depth or generating a final answer from the trajectory of the highest-scoring state, as illustrated in Figure \ref{fig:reasoning_process}. The detailed procedure is presented in Algorithm 2 in Appendix A.

This approach, focusing on maintaining consistency with dialogue-specific knowledge, is complementary to existing paradigms like RAG, which primarily address the integration of external knowledge. The structured nature of D-SMART's reasoning process ensures that the final response is not only contextually relevant but also demonstrably consistent with the facts established throughout the dialogue.

\section{Experimental Setup}
To rigorously evaluate the efficacy of our D-SMART framework, we designed a comprehensive experimental procedure. Our experiments are structured to answer the following research questions (RQs):
\begin{itemize}
    \item \textbf{RQ1:} How effective and stable is D-SMART's performance in multi-turn dialogues compared to state-of-the-art baselines?
    \item \textbf{RQ2:} Can multi-turn consistency be enhanced without compromising response quality?
    \item \textbf{RQ3:} What is the significance of introducing the two core components, DSM and RT, and do they exhibit a synergistic effect? 
\end{itemize}

\subsection{Datasets and Tasks}
We conduct our experiments on MT-Bench-101 \cite{bai-etal-2024-mt}, a benchmark designed for evaluating multi-turn dialogue capabilities. Its structure, featuring 13 task categories that require long-term memory and complex reasoning (e.g., Mathematical Reasoning, Instruction Clarification), provides an ideal testbed for assessing D-SMART's dynamic memory and reasoning abilities. We adopt an autoregressive evaluation setting where models condition on their own previously generated responses. While this may result in lower absolute scores, it stringently tests a model's ability to maintain coherence and recover from its own errors (see Appendix D for task information). 

\subsection{Evaluation Metrics}
Our protocol combines automated quality scoring with novel metrics designed to quantify logical consistency.

\paragraph{GPT Score.}
The holistic quality score (1–10), as judged by GPT-4, is used in MT-Bench-101 to assess overall coherence and task fulfillment. However, while powerful for macro-level assessment, its holistic nature means it can be susceptible to overlooking subtle logical contradictions, particularly when they are masked by a stylistically sound and persuasive response \cite{mendonca-etal-2024-ecoh}.

\paragraph{NLI-based Consistency Metrics.}
To overcome the limitation of GPT score, we introduce two metrics that use a DeBERTa-v3-large model, fine-tuned for Natural Language Inference (NLI), to quantitatively audit for logical integrity and consistency. For each turn, the model computes the probabilities of Entailment ($P_{E_i}$) and Contradiction ($P_{C_i}$) against the dialogue history, providing a targeted measure of a statement's logical grounding.

\begin{itemize}
    \item \textbf{Consistency Score (CS):} Quantifies consistency by rewarding entailment and penalizing contradiction. A score near 1.0 indicates entailment, while a score near 0.0 signals contradiction. The CS of the $i$-th turn of a dialogue can be calculated as:
    \begin{equation}
    \mathrm{CS_i} = \frac{(P_{E_i}-P_{C_i})+1}{2}
    \end{equation}
    
    \item \textbf{Dialogue Entailment Rate (DER):} Measures the proportion of turns classified as ``ENTAILMENT", reflecting the tendency to produce logically grounded responses. It is calculated for a dialogue of $\mathbb{T}$ turns as:
    \begin{equation}
    \mathrm{DER} =  \frac{1}{\mathbb{T}}\sum^{\mathbb{T}}_{i=1} \mathbb{I}(\text{label}_i=\text{ENTAILMENT}) 
    \end{equation}
    where $\mathbb{I}(\cdot)$ is the indicator function. We note that CS is sensitive to the NLI model's confidence (e.g., the margin between $P_{E_i}$ and $P_{C_i}$), while DER is not.
\end{itemize}

\paragraph{Lexical Metrics.}
For completeness, we report Word F1 and BLEU scores, while noting they are not primary indicators of the logical or semantic correctness that our work focuses on.

\subsection{Baselines and Implementation.}
We compare D-SMART against a range of strong models and frameworks. For fair comparison, memory-augmented baselines and D-SMART are built on the same base LLMs.

\paragraph{Proprietary Models.}
\begin{itemize}
    \item \textbf{GPT-4o:} Represents a powerful baseline for models relying on a large (128k) static context window, based on OpenAI's state-of-the-art model.
    \item \textbf{Mem0 (w/ GPT-4o):} A memory framework that uses an LLM to dynamically manage a set of unstructured textual facts extracted from the dialogue.
    \item \textbf{MemoryBank (w/ GPT-4o):} A system utilizes dense retrieval over an expanding long-term memory store, which incorporates a mechanism for memory decay.
\end{itemize}
Both Mem0\cite{chhikara2025mem0} and MemoryBank\cite{10.1609/aaai.v38i17.29946} are representative of memory systems that operate on unstructured text, providing a contrast to D-SMART's structured, formal knowledge graph approach.

\paragraph{Open-Source Models.}
\begin{itemize}
    \item \textbf{Qwen-8B:} A baseline for standard LLM performance without explicit memory modules. Note that its internal ``thinking mode" was disabled in our experiments, which may result in some performance degradation.
    \item \textbf{COMEDY-13B:} An end-to-end framework representing an alternative memory management strategy based on compression of the dialogue history\cite{chen-etal-2025-compress}.
\end{itemize}

\noindent Detailed implementation specifics, hyper-parameters, and a computational cost analysis are provided in Appendix B.

\section{Results and Analysis}
This section presents a comprehensive empirical evaluation structured around our three research questions. Results are averaged over three runs. Detailed case studies are provided in Appendix D.

\begin{table*}[htb]
  \centering
  \begin{threeparttable}
    \small
    \setlength{\tabcolsep}{5pt} 

    \begin{tabular}{llccccc}
      \toprule
      \multicolumn{2}{c}{Model} & GPT Score & CS & DER & F1 & BLEU \\
      \midrule 
      \multirow{4}{*}{GPT-4o} & Cloud-based & 8.20 & 0.594 & 20.94\% & 0.424 & 0.522 \\
      & Mem0 & 8.22 & 0.602 & 21.94\% & \textbf{0.426} & 0.524 \\
      & MemoryBank & 8.30 & 0.621 & 23.88\% & 0.393 & \textbf{0.554} \\
      & \textbf{D-SMART (ours)} & \textbf{8.63} & \textbf{0.692} & \textbf{38.51\%} & 0.414 & 0.549 \\
      \midrule
      COMEDY-13B & & 5.75 & 0.522 & 6.34\% & 0.140 & 0.185 \\
      \cmidrule(lr){2-7}
      \multirow{2}{*}{Qwen3-8B} & Local Inference & 7.79 & 0.627 & 26.23\% & 0.286 & 0.481 \\
      & \textbf{D-SMART (ours)} & \textbf{8.58} & \textbf{0.689} & \textbf{38.73\%} & \textbf{0.388} & \textbf{0.548} \\
      \bottomrule
    \end{tabular}
    \caption{Overall performance comparison on MT-Bench-101. }
    \label{tab:model-comparison}
  \end{threeparttable}
\end{table*}

\subsection{RQ1: Overall Performance against Baselines}
\paragraph{Primary Finding. } 
\textbf{D-SMART significantly enhances both response quality and logical consistency over state-of-the-art baselines} across both proprietary and open-source models. As detailed in Table \ref{tab:model-comparison}, D-SMART applied to GPT-4o achieves a score of 8.63, surpassing the native GPT-4o (8.20) and other memory-augmented methods like Mem0 (8.31). More importantly, it demonstrates superior logical consistency, elevating the DER to 38.51\% from 23.88\% for the next-best baseline (MemoryBank), corresponding to an 84\% improvement over the base model. This is further corroborated by the Consistency Score (CS), which rises to 0.692 from the baseline's 0.594. Notably, this leap in logical integrity does not come at the cost of fluency; the lexical metrics (BLEU, Word F1) remain on par with the strongest baselines, demonstrating that the structured reasoning process does not render the output robotic or unnatural.

\paragraph{Amplifying Open-Source Models. }
The benefits of D-SMART are particularly pronounced on open-source models, where it \textbf{acts as a strong capability enhancer}. When integrated with Qwen-8B, D-SMART improves the score from 7.79 to 8.58 (+10.1\%), a score that approaches the D-SMART-GPT-4o level and far surpasses the much larger COMEDY-13B model (5.75). The consistency metrics show a similar uplift, with DER increasing from 26.23\% to 38.73\% (a 48\% lift over the base model). This result suggests that D-SMART's explicit, structured reasoning can effectively scaffold smaller models, enabling them to approach the performance of larger, state-of-the-art systems.

\begin{figure}[htb]
\centering
\includegraphics[width=0.99\columnwidth]{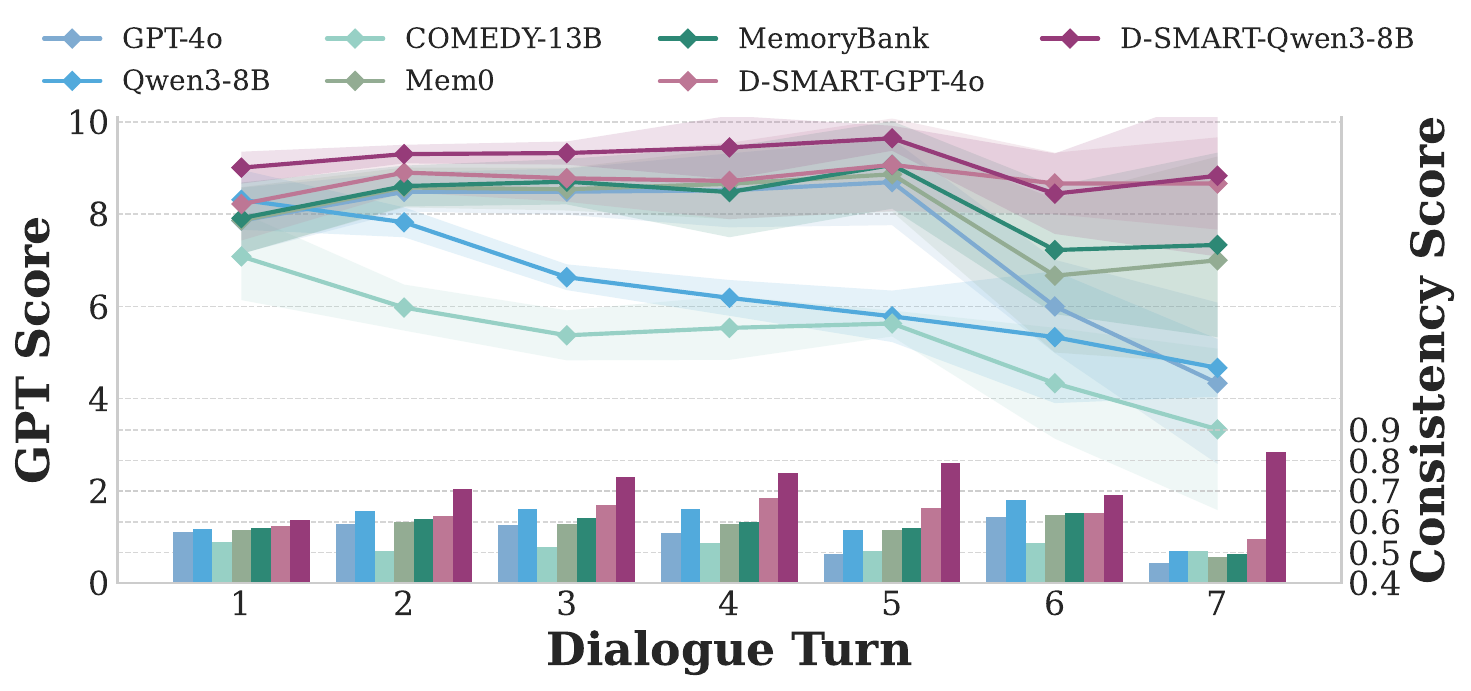} 
\caption{GPT Score (lines) and Consistency Score (bars) across dialogue turns. Error bands represent the standard error of the mean.}
\label{fig:multi_turn}
\end{figure}

\paragraph{Stability in Extended Dialogues. }
D-SMART demonstrates remarkable stability against performance decay in multi-turn long dialogues, a common failure point for existing models. Figure \ref{fig:multi_turn} illustrates that while most baselines, including the powerful GPT-4o and memory-augmented variants, exhibit a sharp decline in performance and consistency in later turns (e.g., after turn 5), both D-SMART-GPT-4o and D-SMART-Qwen-8B maintain high and stable scores throughout the interaction.

The performance collapse in baseline models is intrinsically linked to a decay in logical coherence. As shown by the bars in Figure \ref{fig:multi_turn}, the Consistency Score (CS) of baselines drops in tandem with their GPT score, often falling below the 0.5 threshold, which signifies the generation of self-contradictory content. Conversely, D-SMART sustains the highest Consistency Score (CS) across all turns, demonstrating that \textbf{its structured memory and reasoning process effectively tracks informational shifts to mitigate the logical decay induced by evolving conversational contexts}.

\subsection{RQ2: Balancing Quality and Consistency. }
\begin{figure}[htb]
\centering
\includegraphics[width=0.95\columnwidth]{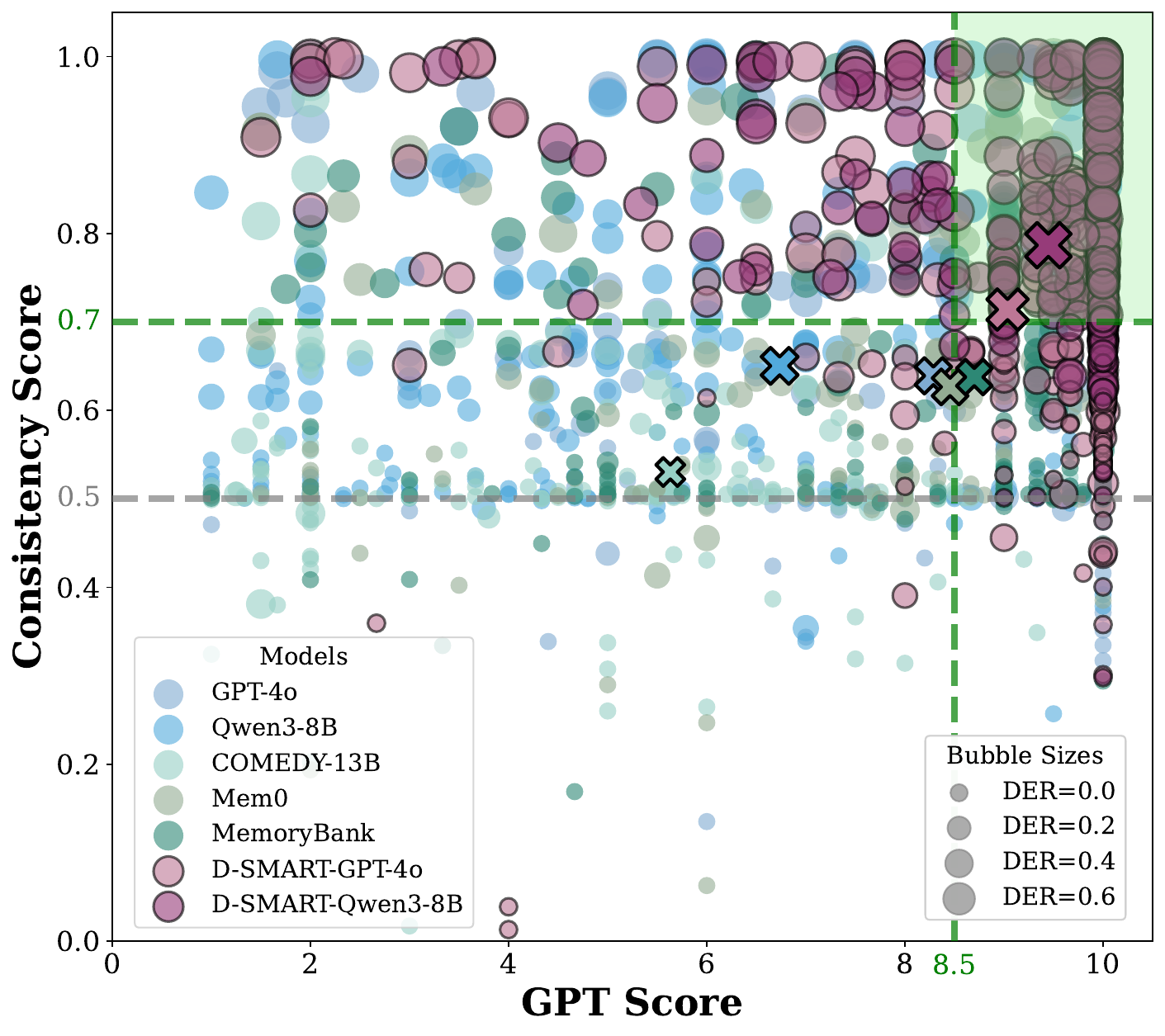} 
\caption{Quality (GPT Score) is plotted against Consistency (CS), with bubble size representing DER. }
\label{fig:bubble}
\end{figure}

A key challenge in dialogue generation is maintaining factual consistency without sacrificing response quality. Figure \ref{fig:bubble} visualizes this trade-off, plotting response quality (GPT Score) against consistency (CS). The results show that baseline models often operate in a ``Risk'' quadrant, producing fluent but logically flawed responses. In contrast, D-SMART models are consistently situated in the ideal upper-right quadrant, demonstrating an \textbf{ability to simultaneously achieve high quality and high consistency}. The average position of D-SMART systems (marked by ``X") is clearly superior on both axes. Furthermore, the larger bubble sizes for D-SMART variants also indicate a higher DER, confirming their outputs are more frequently a direct logical entailment of the dialogue context.

\subsection{RQ3: Ablation Study on Core Components}
\begin{figure}[htb]
  \centering
  \begin{subfigure}[b]{0.48\linewidth}
    \centering
    \includegraphics[width=\columnwidth]{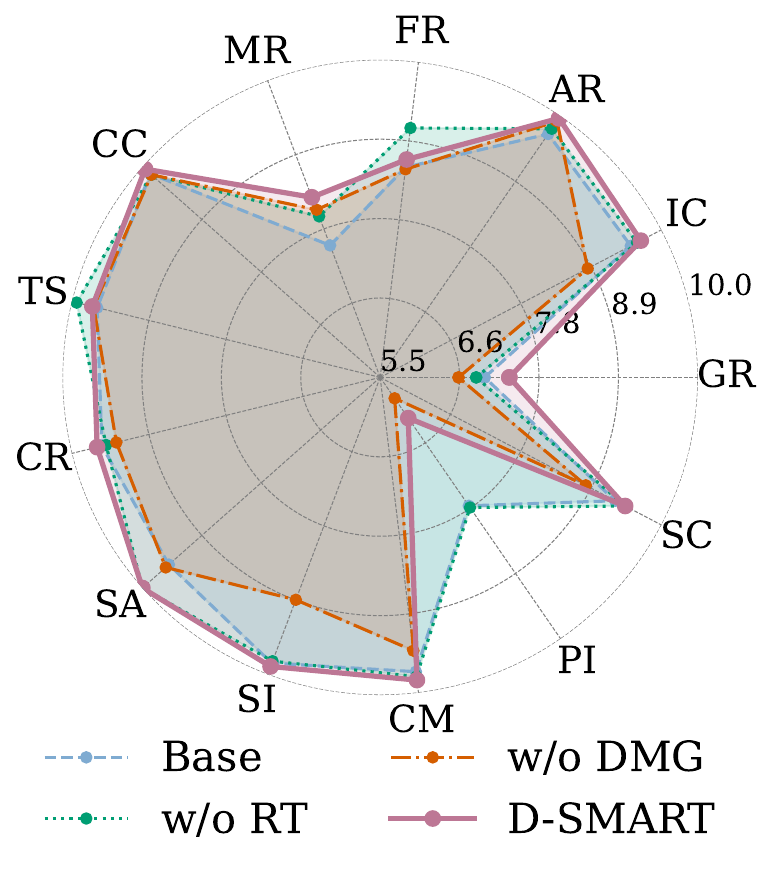}
    \caption{GPT-4o}
    \label{fig:radar_gpt}
  \end{subfigure}
  \hfill
  \begin{subfigure}[b]{0.48\linewidth}
    \centering
    \includegraphics[width=\columnwidth]{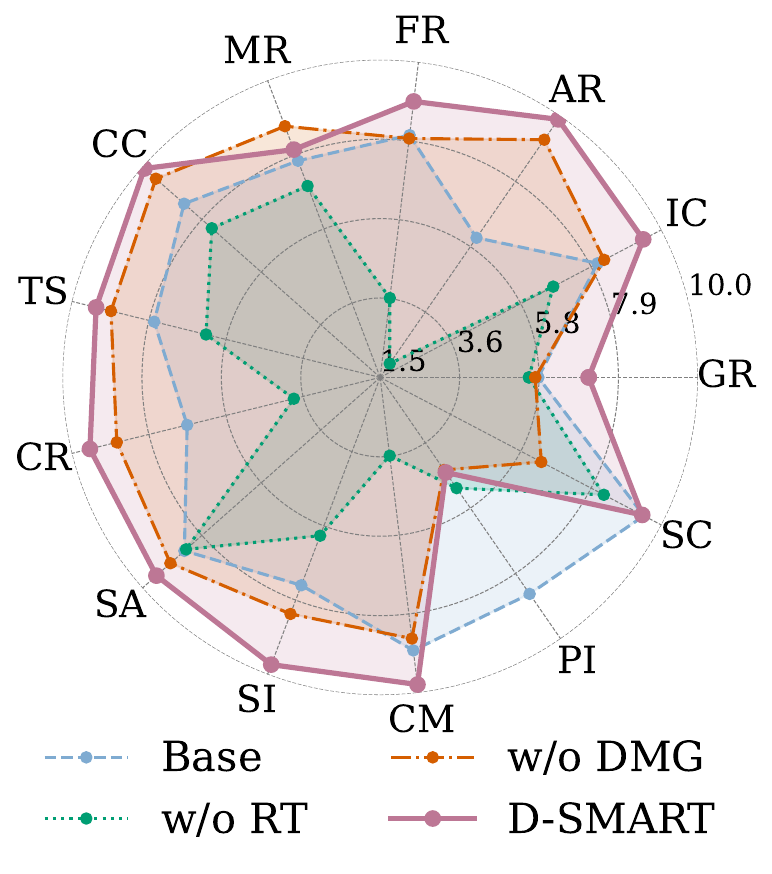}
    \caption{Qwen-8B}
    \label{fig:radar_qwen}
  \end{subfigure}

  \caption{Ablation study on 13 tasks from MT-bench-101 (details in Appendix D) for GPT-4o and Qwen-8B.}
  \label{fig:radar}
\end{figure}

To dissect the interplay between the DSM and the RT, we conducted an ablation study on a curated, more challenging subset of MT-Bench-101. The rationale for this methodological choice is detailed in Appendix~C. The results, presented in Table~\ref{tab:model-comparison-ablation} and Figure~\ref{fig:radar}, reveal a nuanced, model-dependent relationship between the components.

For the highly capable GPT-4o, the DSM-only (``w/o RT") configuration yields the highest GPT Score (9.17), demonstrating the model's intrinsic ability to leverage the factual anchor provided by the DSM. However, its lower consistency scores (CS/DER) compared to the full framework suggest this unguided reasoning lacks discipline. The RT's primary role here is not empowerment but regulation, enforcing a traceable, multi-path reasoning process that remains faithful to the DSM. Conversely, removing the DSM (``w/o DSM") leads to a significant performance drop, as the RT’s structured reasoning becomes ungrounded and speculative.

\begin{table}[htb]
  \centering
  \begin{threeparttable}
    \small
    \setlength{\tabcolsep}{4pt} 
    
    \begin{tabular}{llccc}
      \toprule
      \multicolumn{2}{c}{Model} & GPT Score & CS & DER \\
      \midrule
      \multirow{4}{*}{GPT-4o} 
      & Base &      9.02  & 0.57    &  16.66\%    \\
      & w/o RT &      \textbf{9.17}  & 0.73    &  46.02\%    \\
      & w/o DSM &      8.71  & 0.66    &  33.56\%    \\
      & \textbf{Full (ours)} &      9.11  & \textbf{0.76}    &  \textbf{52.22\%}    \\
      \midrule
      \multirow{4}{*}{Qwen3-8B} 
      & Base &      7.80  & 0.66    &  32.10\%    \\
      & w/o RT &      5.69  & 0.70    &  41.30\%    \\
      & w/o DSM &      7.97  & 0.61    &  23.03\%    \\
      & \textbf{Full (ours)} &      \textbf{8.86}  & \textbf{0.74}    &  \textbf{50.62\%}    \\
      \bottomrule
    \end{tabular}
    \caption{Ablation Analysis of the Impact of D-SMART's Core Components. }
    \label{tab:model-comparison-ablation}
    
  \end{threeparttable}
\end{table}

In contrast, for a smaller model like Qwen-8B, the components are symbiotically indispensable. The DSM-only (``w/o RT") setup results in a severe performance collapse (GPT Score drops to 5.69), indicating the model suffers from "information overload" without the RT acting as a crucial navigator for the knowledge graph. While the RT-only (``w/o DSM") variant provides a minor benefit from structured prompting, the sharp decline in DER (from 32.10\% to 23.03\%) confirms that this reasoning, detached from the DSM's factual basis, is logically inconsistent.

In essence, these findings highlight a clear synergy. \textbf{The DSM provides a dialogue-consistent knowledge base (``what to think about''), while the RT ensures a faithful and deliberate traversal of that base (``how to think'')}. For smaller models, this framework provides an essential cognitive scaffold; for larger models, it refines their powerful capabilities into reliable and verifiable performance.

\section{Related Work}
This section grounds our approach by reviewing three interrelated areas: \textbf{Conversational Memory} for information recall, \textbf{Knowledge Graph Grounding} for factual accuracy, and \textbf{Deliberate Reasoning Frameworks} for structured inference.

\subsection{Conversational Memory in LLMs}
A key research direction involves external memory systems to overcome the limitations of fixed context windows, such as the ``lost in the middle'' problem \citep{liu-etal-2024-lost}. Current approaches include memory compression (e.g., COMEDY \citep{chen-etal-2025-compress}) and retrieval over unstructured text memories (e.g., MemoryBank \citep{10.1609/aaai.v38i17.29946}) to improve information retention. D-SMART is architecturally distinct, employing a formal, OWL-compliant DSM whose semantics enable verifiable consistency checks and deterministic reasoning, capabilities not inherent to text-based memories.

\subsection{Knowledge Graph Grounding for Dialogue}
Another line of work grounds dialogue in Knowledge Graphs (KGs) to improve factual accuracy \citep{chaudhuri2021grounding}. Yet, most approaches leverage large, static KGs and thus cannot adapt to new facts emerging within a conversation \citep{das2018building}. While some research explores dynamic KG construction for specific tasks \citep{lu2025karma, bosselut2021dynamic}, the focus is often on integrating external world knowledge \citep{hixon-etal-2015-learning}. In contrast, D-SMART's DSM is dialogue-specific and dynamic, designed to model and maintain the evolving conversational common ground. This focus necessitates a streamlined mechanism for real-time conflict detection and resolution as dialogue facts are introduced, clarified, or corrected.

\subsection{Deliberate Reasoning Frameworks for LLMs}
To enhance the depth and reliability of LLM inference, various structured reasoning frameworks have been developed. While Chain-of-Thought prompting \citep{wei2022chain} initiated this trend, its strictly linear format restricts exploration and self-correction. The Tree-of-Thoughts (ToT) paradigm \citep{yao2023tree} mitigates these issues by branching multiple reasoning trajectories. For structured inputs, methods such as framing LLMs as KG-traversal agents \citep{sun2023think} or merging LLMs with KGs for graph refinement \citep{pan2024unifying, xu-etal-2024-generate} have emerged. Our RT adopts ToT’s explicit architecture but functions on a symbolic DSM, executing deterministic actions. This shift from unstructured natural-language ``thoughts'' to symbolic traversal enhances control, verifiability, and reliability.

\section{Discussion}
The efficacy of D-SMART is built upon its architectural shift from passive knowledge recall to active, structured reasoning. The framework's dual components, the DSM and RT, synergistically overcome the limitations of relying on static, unstructured dialogue history. The DSM provides a consistently updated, dialogue-specific factual foundation to mitigate context logical decay, while the RT imposes a deliberate, traceable reasoning process upon this memory to guide the model toward sound conclusions. However, the framework's performance is intrinsically linked to the capabilities of the underlying LLM that orchestrates it. The DSM's integrity is contingent on the LLM's proficiency in semantic distillation and conflict adjudication. Similarly, the RT's effectiveness depends on its ability to generate plausible actions and evaluate intermediate states. The advancements demonstrated by D-SMART are thus predicated on the base model's semantic and logical competence. Understanding and enhancing the LLM's reliability in these fine-grained control tasks is therefore a crucial direction for maximizing the framework's potential and for future research.

A primary limitation of this approach is the increased computational overhead and response latency as RT extends LLM inference time by branching multiple reasoning paths, our experiments show that the average inference time per turn for a local open-source model increases from approximately 0.3s to 1.3s. The subsequent memory maintenance requires approximately 6s per turn (detailed computational cost analysis are provided in Appendix B); note that it can be performed asynchronously, allowing it to overlap with the user's input time, thereby mitigating its impact on the interactive flow. We contend this is a deliberate and necessary architectural trade-off. While methods relying on ever-larger context windows are faster, they remain susceptible to factual inconsistency and the ``lost in the middle" problem in extended dialogues. D-SMART, in contrast, exchanges a measure of speed for a significant enhancement in logical coherence and stability, achieving a near 48\% increase in dialogue consistency (DER) for open-source models. This positions the framework as a viable option for applications where reliability and trustworthiness are paramount.

\section{Conclusion and Future Work}
In this paper, we proposed D-SMART, a model-agnostic framework that enhances dialogue consistency by coupling a Dynamic Structured Memory with a Reasoning Tree. Our experiments show that by grounding the LLM's generation process in an explicit, dynamically updated knowledge base, D-SMART significantly mitigates logical decay in multi-turn dialogues and improves response quality, particularly in extended dialogues where baseline models falter.

Future work will proceed along three main directions: enhancing the DSM with more sophisticated memory management, optimizing the RT's search efficiency to reduce latency through techniques such as heuristic pruning and batched parallel generation, and expanding the framework's scope to integrate external knowledge.

\bibliography{main}

\newpage
\appendix
\section{Appendix A: Algorithm Descriptions and Pseudocode}
\subsection{KGE Pipeline for Knowledge Graph Generation}
The KGE pipeline is a neuro-symbolic process that transforms a natural language statement $s_t$ from a dialogue turn into a formal, OWL-compliant knowledge graph fragment $\mathcal{G}'_{t}$.   This is a critical step in the Memory Maintenance Phase.   The pipeline first parses the text into an Abstract Meaning Representation (AMR) to capture its core semantic structure.   This AMR is then converted into a preliminary OWL graph.   Finally, a semantic enrichment stage enhances the graph's logical rigor by linking ambiguous entities to formal lexical resources and high-level ontologies using Word Sense Disambiguation (WSD).   The resulting grounded fragment is then ready for integration into the main DSM.

\begin{algorithm}[htb]
\caption{KGE Pipeline for Knowledge Graph Generation}
\label{alg:kge}
\textbf{Input:} Text $T$ \\
\textbf{Output:} RDF/OWL Knowledge Graph $\mathcal{G}$

\begin{algorithmic}[1]
\STATE \COMMENT{Module 1: Text to Abstract Meaning Representation Parsing}
\STATE \COMMENT{Parse text into AMR form}
\STATE $\mathcal{G}_{AMR} \leftarrow \text{SPRING}(T)$ 

\STATE \COMMENT{Module 2: AMR to OWL KG Translation}
\STATE \COMMENT{Transform AMR to an initial OWL-compliant RDF KG }
\STATE $\mathcal{G} \leftarrow \text{AMR2FRED}(\mathcal{G}_{AMR})$ 

\STATE \COMMENT{Semantic Enrichment}
\FOR{each node $n$ in $\mathcal{G}_{AMR}$ that lacks a link to a lexical resource}
    \STATE \COMMENT{Perform Word Sense Disambiguation (WSD)}
    \STATE \COMMENT{Get word synset from Framester via EWISER }
    \STATE $S_{synset} \leftarrow \text{EWISER}(T, \text{lemma}(n))$ 
    \STATE $\mathcal{G} \leftarrow \mathcal{G} \cup \{ (n, \text{owl:equivalentClass}, S_{synset}) \}$

    \STATE \COMMENT{Align to top-level ontologies}
    \STATE \COMMENT{Get word ``supersense" }
    \STATE $S_{supersense} \leftarrow \text{Framester.getSupersense}(S_{synset})$ 
    \STATE $\mathcal{G} \leftarrow \mathcal{G} \cup \{ (n, \text{rdfs:subClassOf}, S_{supersense}) \}$

    \STATE \COMMENT{Get DOLCE+DnS Ultra Lite (DUL) class }
    \STATE $C_{DUL} \leftarrow \text{Framester.getDULClass}(S_{synset})$ 
    \STATE $\mathcal{G} \leftarrow \mathcal{G} \cup \{ (n, \text{rdfs:subClassOf}, C_{DUL}) \}$
\ENDFOR

\STATE \textbf{return} $\mathcal{G}$
\end{algorithmic}
\end{algorithm}

\subsection{Beam Search for Reasoning Tree Traversal}
\label{sec:reasoning_algo}
Algorithm~\ref{alg:beam_search} operationalizes the Reasoning Tree (RT) traversal, employing a beam search to find a grounded response $r_t$ for a query $q_t$ by reasoning over the memory graph $\mathcal{G}_{t-1}$. The search iteratively expands a tree of reasoning paths. At each step, for the most promising states on the beam, an LLM acting as a policy ($p_{\theta}^{\pi}$) proposes actions (e.g., \texttt{EXPAND\_ENTITY}, \texttt{Think}). Actions involving graph traversal are executed deterministically against the DSM to ensure factual grounding. An LLM acting as a value function ($p_{\theta}^{v}$) then scores the newly generated states. The beam is subsequently pruned to retain only the top-$k$ candidates, which balances exploration of multiple reasoning paths with computational tractability. The process terminates upon generating a final answer or reaching a maximum depth, with the response extracted from the highest-scoring trajectory.

\begin{algorithm}[htb]
\caption{Beam Search for Reasoning Tree Traversal}
\label{alg:beam_search}
\textbf{Input:} Query $q_t$, Dynamic Structured Memory $\mathcal{G}_{t-1}$, beam width $k$, max depth $D_{max}$ \\
\textbf{Output:} Final response $r_t$

\begin{algorithmic}[1]
\STATE Initialize root state $\mathcal{S}_0 \leftarrow (\tilde{\mathcal{G}}_0=\emptyset, \mathcal{Z}_0=\langle\rangle, v_0=1.0, d_0=0)$
\STATE Initialize beam $B \leftarrow \{\mathcal{S}_0\}$
\FOR{$d = 0$ to $D_{max}-1$}
    \STATE Initialize candidate pool $C \leftarrow \emptyset$
    \FOR{each state $\mathcal{S}_i$ in $B$}
        \IF{$\mathcal{S}_i$ is not a terminal state}
            \STATE Propose actions $A_{cand} \sim p_\theta^\pi(a \mid q_t, \mathcal{S}_i)$
            \FOR{each action $a_{ij}$ in $A_{cand}$}
                \STATE Execute $a_{ij}$ on $\mathcal{S}_i$ to get a new state $\mathcal{S}_j$
                \STATE \COMMENT{e.g., query $\mathcal{G}_{t-1}$ for $a_{ee}, a_{fp}$}
                \STATE Evaluate new state: $v_j \sim p_\theta^{v}(v \mid q_t, \mathcal{S}_j)$
                \STATE Add $\mathcal{S}_j$ to $C$
            \ENDFOR
        \ELSE
            \STATE Add $\mathcal{S}_i$ to $C$ \COMMENT{Keep terminal states}
        \ENDIF
    \ENDFOR
    \STATE Sort $C$ by value scores $v_j$ in descending order
    \STATE Update beam $B \leftarrow \text{top-}k \text{ states from } C$
    \IF{all states in $B$ are terminal}
        \STATE \textbf{break}
    \ENDIF
\ENDFOR
\STATE Select final state $\mathcal{S}^* = \underset{\mathcal{S} \in B}{\mathrm{argmax}} \ v$
\STATE Extract response $r_t$ from trajectory $\mathcal{Z}^*$ of $\mathcal{S}^*$
\STATE \textbf{return} $r_t$
\end{algorithmic}
\end{algorithm}

\clearpage
\section{Appendix B: Implementation and Computational Cost Analysis}
\label{sec:appendix_impl_details}
To ensure the transparency and reproducibility of our research, this section provides a detailed account of our experimental setup, including the hardware and software environment, hyperparameter configurations for the D-SMART framework, implementation details for baseline models, and a comprehensive analysis of computational costs.

\subsection{Environment and Dependencies}
All experiments were conducted on the Ubuntu 22.04 operating system. Our framework is implemented using PyTorch and the HuggingFace Transformers library for model handling. The DSM is managed and maintained using a suite of specialized libraries for knowledge graph operations, including RDFlib, NetworkX, owlrl, and owlready2. For proprietary models, we utilized the official OpenAI API. The specific versions of all libraries are detailed in the provided code repository.

The hardware infrastructure was distributed as follows to optimize our experimental workflow:
\begin{itemize}
    \item \textbf{Local Model Execution:} All open-source models (e.g., Qwen-8B, COMEDY-13B) were run on a server equipped with an NVIDIA H800 GPU, a 20-core Intel(R) Xeon(R) Platinum 8458P CPU, and 100GB of RAM.
    \item \textbf{API-based Model Execution:} Experiments involving API calls (e.g., GPT-4o, Mem0, MemoryBank) were conducted on a machine with an NVIDIA GeForce RTX 4090 GPU, a 32-core 13th Gen Intel(R) Core(TM) i9-13900K CPU, and 128GB of RAM.
    \item \textbf{Evaluation:} All evaluation processes, including the execution of NLI-based metrics, were performed on a workstation with an NVIDIA GeForce RTX 3080 GPU, a 20-core Intel(R) Core(TM) i9-10900K CPU, and 96GB of RAM.
\end{itemize}
To ensure full reproducibility, we will release our full code and experimental configurations at Github.

\subsection{Baseline and Framework Implementation Details}

\paragraph{D-SMART Hyperparameter Configuration. }
The performance of the D-SMART framework is influenced by the parameters governing the RT search process. For all main experiments reported in this paper, we adopted the following configuration, which was determined through a sensitivity analysis detailed below:
\begin{itemize}
    \item \textbf{Beam Search Width ($B$):} 3
    \item \textbf{Action Samples per Node ($S$):} 3
    \item \textbf{Maximum Search Depth ($D_{max}$):} 5
    \item \textbf{Maximum Expansion Steps ($M_{exp}$):} 12
\end{itemize}

\paragraph{Parameter Sensitivity Analysis.}
To understand the impact of key hyperparameters on performance, we conducted a sensitivity analysis on a randomly sampled subset of 100 dialogues from MT-Bench-101. We designed seven distinct configurations to systematically evaluate how reasoning depth ($D_{max}$), beam search width ($B$), and the number of action samples ($S$) affect both dialogue consistency (Consistency Score) and overall quality (GPT Score).

\begin{figure}[htb]
\centering
\includegraphics[width=0.99\columnwidth]{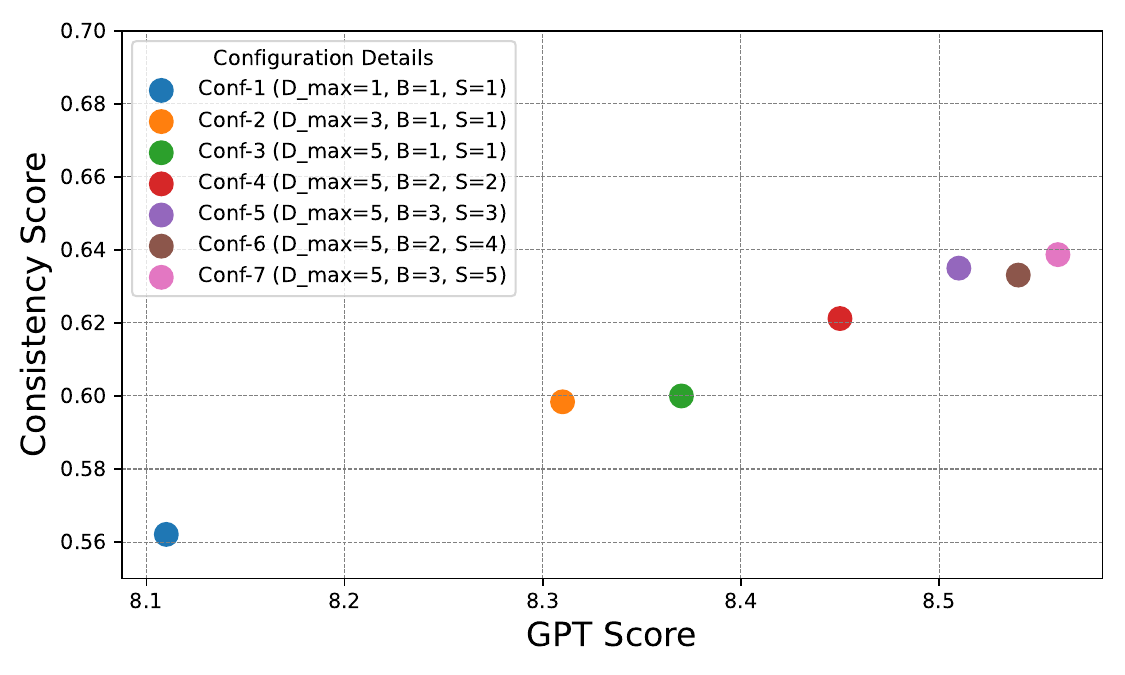} 
\caption{GPT Score and Consistency Score under different hyperparameters.}
\label{fig:para_conf}
\end{figure}

The results, as illustrated in Figure~\ref{fig:para_conf}, shows that increasing the reasoning depth ($D_{max}$ from 1 to 5, cf. Conf-1, 2, 3) consistently improves both quality and consistency. A deeper search allows the framework to form more complex reasoning chains. We observed an interesting pattern where the framework often uses a \texttt{THINK} action to generate a preliminary answer draft before the final \texttt{ANSWER} action. This two-step process—first establishing a logically sound core, then refining its linguistic expression—appears to foster responses that are both coherent and articulate.

Similarly, widening the beam search ($B$ from 1 to 3, cf. Conf-3 vs. Conf-5) yields significant gains. A wider beam enables the RT to explore multiple reasoning paths in parallel. Some paths might prioritize linguistic fluency, while others may ensure factual grounding through more frequent interactions with the DSM (e.g., via \texttt{EXPAND\_ENTITY}). The beam search mechanism allows the system to systematically compare these trajectories and select the one that is most logically consistent and grounded in the established dialogue facts, mitigating the risk of contradiction inherent in single-threaded reasoning.

Based on this analysis, we selected \textbf{Conf-5 ($D_{max}=5, B=3, S=3$)} as our final configuration for all main experiments. This setting achieves a strong balance between high performance and manageable computational overhead.

\paragraph{Large Language Model (LLM) API Parameters.}
To ensure deterministic and reproducible results where necessary, we carefully controlled the LLM API parameters.
\begin{itemize}
    \item \textbf{Temperature:} For the state evaluation step within the RT, where deterministic scoring is crucial, the temperature was set to $0.0$. For all other generation tasks, a temperature of $0.8$ was used to encourage more diverse and natural-sounding responses.
    \item \textbf{Max Tokens:} The maximum number of generated tokens was set to $8192$.
    \item \textbf{Model Versions:} We used the specific model versions \texttt{gpt-4o-2024-05-13} and \texttt{qwen3-8b} to ensure that our results can be precisely replicated in future studies. All other parameters were left at their default settings.
\end{itemize}

\paragraph{Baseline Implementation.}
We used the official code implementations for all baseline models and frameworks to ensure a fair and accurate comparison.
\begin{itemize}
    \item \textbf{Mem0 \& MemoryBank:} We adapted the official code by replacing the original backbone LLM with \texttt{GPT-4o}, aligning it with our D-SMART setup. Key internal parameters were kept at the default values recommended in their respective papers. Our observations indicate that Mem0's reliance on the LLM for single-step decisions over unstructured text limits its ability to maintain long-term consistency with complex or contradictory information. MemoryBank's memory update mechanism, inspired by the Ebbinghaus forgetting curve, is driven by temporal relevance rather than logical consistency, which can lead to the decay of crucial facts over extended dialogues.
    \item \textbf{COMEDY-13B:} We utilized the officially released model weights and inference scripts. As an end-to-end framework, its compressive memory mechanism, while efficient, was observed to occasionally discard critical numerical or entity details during the compression process, which subsequently impaired its logical consistency on tasks requiring high precision.
\end{itemize}

\subsection{Computational Cost Analysis}

\paragraph{Algorithmic Complexity Analysis.}
The computational overhead of D-SMART stems primarily from two phases: the Memory Maintenance Phase and the Response Phase. Let the key variables be defined as follows: $N$ is the total number of triples in the DSM, $N'$ is the number of triples in a newly extracted graph fragment, $V$ and $E$ are the number of nodes and edges in the DSM, $T_{llm}$ is the average latency of a single LLM API call, $M_{exp}$ is the maximum expansion steps in the RT, $B$ is the beam width, and $S$ is the number of action samples.

\begin{itemize}
    \item \textbf{Memory Maintenance Phase:} This phase involves knowledge extraction and graph updates.
    \begin{itemize}
        \item \textit{Knowledge Fragment Extraction:} The complexity is dominated by the KGE pipeline call, which we denote as $O(T_{kge})$.
        \item \textit{Conflict Detection:} This relies on an LLM call to compare graph serializations. The complexity is approximately $O(N + N' + T_{llm})$, where the LLM call is the main bottleneck.
        \item \textit{Graph Merging:} This involves set operations on the graph triples, with a complexity of roughly $O(N + N')$.
    \end{itemize}
    
    \item \textbf{Response Phase (Reasoning Tree Search):} The complexity is governed by the beam search algorithm. The total complexity can be approximated as:
    \begin{equation}
    \begin{split}
    \text{Complexity} \approx M_{exp} \times \Big[ & O(N_{tree} \times \log N_{tree}) \\
    & + B \times S \times (2T_{llm} \\
    & + O(V+E)) \Big]
    \end{split}
    \end{equation}
    where $N_{tree}$ is the number of nodes in the search tree. The term $O(V+E)$ arises from the most expensive graph traversal action, \texttt{FIND\_PATH}, which invokes a shortest-path search on the entire graph. The $2T_{llm}$ term accounts for the two LLM calls per expansion: one for action sampling and one for state evaluation. The parameters $M_{exp}$, $B$, and $S$ are the primary levers for balancing performance and computational cost. We also note that the number of effective actions generated by the LLM is often less than or equal to $S$, as duplicate actions may be proposed, slightly reducing the effective branching factor.
\end{itemize}

\paragraph{Empirical Latency Measurement.}
We measured the average end-to-end inference latency per dialogue turn on an H800 GPU and an 4090 GPU to provide a practical assessment of the overhead, as shown in Table~\ref{tab:latency_comparison}.

\begin{table}[htb]
\centering
\begin{tabular}{lrr}
\toprule
\textbf{Model} & \textbf{Baseline (s)} & \textbf{D-SMART (s)} \\
\midrule
Qwen3-8B & 0.27 & 1.27 \\
GPT-4o API & 3.50 & 9.80 \\
\bottomrule
\end{tabular}
\caption{Average per-turn inference latency comparison. API-based models include network latency.}
\label{tab:latency_comparison}
\end{table}

The overhead introduced by D-SMART is significant but represents a deliberate architectural trade-off for enhanced logical consistency and reliability. The latency increase in the Response Phase is primarily due to the multiple LLM calls for action sampling and state evaluation in the RT.

The Memory Maintenance Phase, while computationally intensive, can be optimized. Its average latency is approximately 6 seconds, broken down as follows:
\begin{itemize}
    \item Structured Statement Generation: $\sim$0.13s
    \item Knowledge Fragment Extraction (KGE API): $\sim$5s
    \item Conflict Detection \& Merging: $\sim$0.75s
\end{itemize}
Note that this entire maintenance phase can be executed \textbf{asynchronously}. It can process the completed turn $(q_t, r_t)$ while the user is preparing their next query $q_{t+1}$, thereby hiding most of this latency from the user and minimizing its impact on the interactive flow of the dialogue. Future work can further mitigate costs by developing more efficient search heuristics for the RT and employing smaller, specialized models for internal tasks like state evaluation.

\clearpage
\section{Appendix C: Details on the Ablation Study}
\label{sec:appendix_ablation}
\subsection{Rationale for Using a Refined Subset for Ablation Analysis}
To rigorously dissect the individual contributions and the synergistic interplay of our core components, the DSM and the RT, we conducted the ablation study on a strategically curated subset of the MT-Bench-101 benchmark. A key motivation for this approach is to mitigate the confounding effects of performance saturation observed when evaluating on the full dataset.

Our preliminary analysis revealed that a significant portion of the MT-Bench-101 dialogues falls into two categories that limit the conclusiveness of ablation results:
\begin{itemize}
    \item \textbf{Ceiling Effect:} Many tasks are simple enough that even baseline models can achieve high scores \cite{10.5555/3454287.3454581}. In these scenarios, the performance is already saturated, leaving little to no room to measure the true uplift provided by components like DSM and RT.
    \item \textbf{Floor Effect:} Conversely, some dialogues are exceedingly difficult or ill-posed, causing even the most advanced models to fail. In these cases, the models operate in a near-random-guess regime, and the impact of our framework's components is masked by the fundamental difficulty of the task.
\end{itemize}

These effects can lead to an ``Average Effect Masking,'' where the true, significant impact of our components on moderately to highly complex tasks is diluted by the large number of non-informative cases. Therefore, to ensure a sensitive and meaningful evaluation, we designed a principled methodology to isolate a subset of dialogues that are both challenging and solvable, thus providing a clearer window into the components' efficacy.

\subsection{Selection Methodology and Criteria}
We curated this refined subset, which constitutes 25\% of the original MT-Bench-101 dataset, through a two-stage, criteria-driven filtering process designed to maximize task complexity and discriminative power.

\textbf{Stage 1: Performance-based Filtering.} To eliminate dialogues susceptible to ceiling or floor effects, we first filtered the dataset based on the performance of the \texttt{Qwen-8B} baseline model. We retained only those dialogues where the baseline model achieved a GPT score in a moderate range (e.g., between 4.0 and 8.0). This step effectively removes tasks that are either too trivial or prohibitively difficult for a standard model, ensuring our analysis focuses on dialogues where improvements are both possible and measurable.

\textbf{Stage 2: Complexity-based Curation.} From the pool of dialogues selected in Stage 1, we then ranked them based on a suite of linguistic and structural complexity metrics, similar to how \citeauthor{perez-rosas-etal-2019-makes} do. The goal was to identify dialogues that place high demands on memory, reasoning, and context management. The primary criteria for this ranking included:
\begin{itemize}
    \item \textbf{Dialogue Scale and Information Density:} A high \textit{Average Turns per Dialogue} and \textit{Average Words per Turn} to test the model's long-term memory and its ability to process information-dense utterances, which are core challenges for the DSM.
    \item \textbf{Semantic Richness and Entity Management:} A high \textit{Vocabulary Diversity (TTR)} and \textit{Average Unique Entities} per dialogue. These metrics identify conversations with broad topics and numerous entities, directly challenging the DSM's capacity for knowledge construction and maintenance.
    \item \textbf{Logical and Structural Complexity:} A high \textit{Conditional Statement Frequency} and \textit{Negative Constraint Frequency}. These metrics were chosen to specifically probe the model's logical reasoning capabilities and its adherence to complex constraints, functions directly supported by the RT framework.
    \item \textbf{Context Dynamics:} A high \textit{New Entity Introduction Rate}, which signals a rapidly evolving conversational context. This tests the DSM's dynamic update mechanism, including its ability to integrate new facts and resolve conflicts in real-time.
\end{itemize}

We selected the top-ranked dialogues from this process, finally we get a subset reached 25\% of the original dataset's size, we ensured the final set for the ablation study was composed of the most challenging and informative examples.

\subsection{Validation of the Refined Subset}
To empirically validate the outcome of our selection methodology, we performed a statistical comparison between the curated subset and the original MT-Bench-101 dataset. As detailed in Table~\ref{tab:ablation_subset_stats}, the refined subset demonstrates significantly higher complexity across all measured dimensions. For instance, the average number of unique entities per dialogue nearly doubled (from 5.04 to 9.91), and the introduction rate of new entities per turn increased by over 74\% (from 1.66 to 2.89). This confirms that our principled approach successfully isolated a more challenging testbed, making it ideal for a nuanced and insightful ablation analysis.
\begin{table}[htb]
\centering
\begin{tabular}{lrr}
\toprule
\textbf{Metric} & \textbf{Original} & \textbf{Refined} \\
\midrule
Avg. Turns per Dialogue & 3.03 & 3.42 \\
Avg. Words per Turn & 66.3 & 82.99 \\
Vocabulary Diversity (TTR) & 6.12\% & 10.89\% \\
Avg. Unique Entities & 5.04 & 9.91 \\
Conditional Freq. & 16.96\% & 22.46\% \\
Negative Constraint Freq. & 12.36\% & 18.19\% \\
New Entity Intro. Rate & 1.66 & 2.89 \\
\bottomrule
\end{tabular}
\caption{Statistical comparison between the original MT-Bench-101 dataset and the refined, more challenging subset used for the ablation study. The refined subset shows consistently higher complexity across all metrics.}
\label{tab:ablation_subset_stats}
\end{table}

\clearpage
\onecolumn
\section{Appendix D: Benchmark Task Details and Case Studies}
The MT-Bench-101 benchmark is structured around 13 distinct task categories, each formulated to rigorously evaluate specific facets of a model's multi-turn dialogue capabilities, such as logical consistency and the faithful recall of long-term memory. To offer both a formal definition and a concrete understanding of the challenges posed by this benchmark, we first present a table detailing the full name, abbreviation, and description for each of the 13 tasks. Following this, we present dialogue examples (case studies) from key categories to illustrate how these tasks probe for specific capabilities in practice.

Note that in each case, \colorbox{ProcessBlue!40}{\textcolor{black}{\textbf{light blue}}} indicates information the model will recall in the future, \colorbox{DarkSeaGreen!60}{\textcolor{black}{\textbf{light green}}} indicates the model’s correct responses, \colorbox{red!20}{\textcolor{black}{\textbf{light red}}} indicates the model’s incorrect responses, and \colorbox{orange!40}{\textcolor{black}{\textbf{light orange}}} indicates the contradictory or inconsistent content in the context. Each model or framework has “\encircle{E}” on the left indicating that the NLI relation for that turn with the context was judged as \textbf{ENTAILMENT}, “\encircle{N}” indicates \textbf{NEUTRAL}, and “\encircle{C}” indicates \textbf{CONTRADICTION}.

\subsection{MT-Bench-101 Task Descriptions}
Detailed task description of MT-Bench-101 are shown in Table~\ref{tab:mt-bench-101-tasks}.
\begin{table*}[htb]
\centering
\small
\begin{tabular}{p{2.5cm} p{1.5cm} p{11cm}}
\toprule
\textbf{Task} & \textbf{Abbr.} & \textbf{Task Description} \\
\midrule
Context Memory & CM & Evaluates the model's ability to recall details from earlier in the dialogue to address a user's current question, assessing the consistency of its long-term memory and contextual continuity. \\
\hline
Anaphora Resolution & AR & Assesses the model's capacity to correctly resolve anaphoric references (e.g., pronouns like ``it",  ``these") throughout a conversation, which is crucial for maintaining semantic consistency. \\
\hline
Separate Input & SI & Measures the model's ability to maintain a consistent link between task instructions from an initial turn and the specific inputs provided in subsequent turns, a key test of operational consistency. \\
\hline
Topic Shift & TS & Tests the model's agility in recognizing an abrupt topic switch from the user and maintaining conversational coherence by focusing on the new topic without being unduly influenced by prior context. \\
\hline
Content Confusion & CC & Probes the model's skill in distinguishing between textually similar queries that have different meanings, testing its ability to avoid interference and maintain logical consistency in its responses. \\
\hline
Content Rephrasing & CR & Requires the model to rephrase its previous response based on a new user requirement (e.g., simplification). This tests its ability to maintain informational consistency while altering the presentation. \\
\hline
Format Rephrasing & FR & Involves transforming the structure of a previous response (e.g., from paragraph to list) while preserving the original information, assessing its ability to maintain content fidelity across formats. \\
\hline
Self-correction & SC & Assesses the model's ability to revise an erroneous response after receiving valid user feedback, thereby restoring consistency with factual accuracy. \\
\hline
Self-affirmation & SA & Evaluates the model's capacity to adhere to its correct response when faced with an incorrect user challenge, a test of its confidence and consistency of knowledge. \\
\hline
Mathematical Reasoning & MR & Tests collaborative problem-solving where the model must maintain logical consistency while incrementally incorporating new conditions across turns to solve a complex mathematical problem. \\
\hline
General Reasoning & GR & Assesses the ability to work with users over successive turns to solve complex reasoning puzzles, which requires maintaining a consistent line of logical deduction or induction. \\
\hline
Instruction Clarification & IC & Targets the model's ability to handle ambiguous queries by asking clarifying questions, a strategy to ensure the subsequent dialogue is consistent with the user's true intent. \\
\hline
Proactive Interaction & PI & Measures the ability to pose relevant questions in response to user statements, aiming to maintain a coherent and engaging dialogue rather than letting the conversation terminate prematurely. \\
\bottomrule
\end{tabular}
\caption{Detailed descriptions of the 13 tasks within MT-Bench-101.}
\label{tab:mt-bench-101-tasks}
\end{table*}

\clearpage
\subsection{Case Studies}
\paragraph{Case.1: } This case study illustrates the difference in \textbf{factural consistency} between D-SMART and a representative baseline framework (MemoryBank) when tasked with rephrasing and reformatting information across multiple turns. While both frameworks produce high-quality, fluent responses, MemoryBank introduces a subtle but significant \textbf{logical conflict}, a failure that D-SMART’s architecture is designed to prevent. Initially, both frameworks identify several risks of AI. D-SMART grounds its subsequent responses in the seven points it first established, maintaining informational integrity, as reflected by its high Consistency Scores (CS). In contrast, MemoryBank demonstrates a \textbf{contextual inconsistency}. In its first response, ``developers" are framed as part of a collective group responsible for addressing AI risks. However, in its final turn, it introduces new, un-grounded information by stating that AI systems can reflect ``the biases of their developers", positioning developers as a source of the problem. This shift creates a conflict with the previously established information. This inconsistency is captured by its low CS of 0.57, even as it receives a high GPT Score (GS) of 10 for its fluency and surface-level correctness. 
This example highlights how D-SMART ensures that \textbf{responses remain factural consistent}.  Baselines lacking such a mechanism may fail to preserve consistency, especially in nuanced aspects of the dialogue history, a flaw that can be \textbf{overlooked by holistic quality metrics} like the GPT Score.

\begin{case}[label={ex:case_1}]{Case.1: Maintaining Factual Consistency Across Dialogue Turns}
\tiny

{\tiny \textbf{Turn 1}}

User: What are the potential risks of artificial intelligence?

\noindent\rule{\linewidth}{0.4pt}

Assistant:
\begin{tcbitemize}[
    raster columns=2,
    raster equal height,
    colframe=yellow!20!white,         
    colback=yellow!20!white,          
    colbacktitle=yellow!20!white,     
    coltitle=red!60!black, 
    fonttitle=\bfseries,
    after title={\vspace{-10pt}},
    before skip=0pt,
    after skip=0pt
    ]
\tcbitem[title={\scriptsize \encircle{N} D-SMART \text{$(\textrm{CS}_1=0.69)$}}, coltitle=green!50!black]
    Artificial intelligence (AI) has several potential risks, including:
    
    1. \textbf{Job Loss}: Automation through AI could replace human workers in various industries, leading to unemployment and economic challenges.
    
    2. \textbf{Bias and Discrimination}: AI systems can inherit biases from the data they are trained on, leading to unfair or discriminatory decisions.
    
    3. \textbf{Privacy Concerns}: AI-powered technologies can collect and analyze vast amounts of personal data, raising concerns about surveillance and misuse of information.
    
    4. \textbf{Security Risks}: AI can be exploited for cyberattacks, creating vulnerabilities in systems and potentially endangering critical infrastructure.
    
    5. \textbf{Unintended Consequences}: Poorly designed AI systems may behave unpredictably, causing harm or making irresponsible decisions.
    
    6. \textbf{Ethical and Moral Issues}: AI raises questions about accountability for actions, especially in areas like autonomous weapons or life-altering decisions.
    
    7. \textbf{Concentration of Power}: The development of AI could lead to increased power for a few companies or nations, potentially creating inequality and dependency.

\tcbitem[title={\scriptsize \encircle{N} MemoryBank \text{$(\textrm{CS}_1=0.51)$}}, coltitle=red!70!black]
    Artificial intelligence has the potential to bring about many benefits, such as improving efficiency and advancing technological developments. However, there are also some risks associated with AI. Some concerns include the possibility of job displacement due to automation, the ethical implications of AI decision-making, privacy issues related to data collection and surveillance, and the potential for bias in AI algorithms. \colorbox{orange!40}{\textcolor{black}{It is important for researchers, developers, policymakers, and society as a whole}} to work together to address these risks and ensure that AI is developed and deployed responsibly.
\end{tcbitemize}

\noindent\rule{\linewidth}{0.4pt}

{\tiny \textbf{Turn 2}}

User: Can you rephrase your answer in simpler terms, focusing on the main ideas?

\noindent\rule{\linewidth}{0.4pt}

Assistant:
\begin{tcbitemize}[
    raster columns=2,
    raster equal height,
    colframe=yellow!20!white,         
    colback=yellow!20!white,          
    colbacktitle=yellow!20!white,     
    coltitle=red!60!black, 
    fonttitle=\bfseries,
    after title={\vspace{-10pt}},
    before skip=0pt,
    after skip=0pt,
    ]
\tcbitem[title={\scriptsize \encircle{E} D-SMART \text{$(\textrm{GS}_2=10, \textrm{CS}_2=0.98)$}}, coltitle=green!50!black]
    AI can cause problems like replacing jobs, making unfair decisions due to biased training data, invading privacy by collecting too much personal information, being used in cyberattacks, acting in unpredictable ways, and raising ethical questions about accountability. It can also create inequalities if only a few powerful companies or countries control and benefit from it.

\tcbitem[title={\scriptsize \encircle{E} MemoryBank \text{$(\textrm{GS}_2=8, \textrm{CS}_2=0.72)$}}, coltitle=red!70!black]
    Sure! Artificial intelligence can bring good things, like making things easier and creating new technology. But it can also have some problems. For example, it might take away people's jobs, make decisions that aren't fair, invade privacy by collecting too much data, or be biased. It's important for everyone to work together to make sure AI is used in a good and fair way.
\end{tcbitemize}

\noindent\rule{\linewidth}{0.4pt}

{\tiny \textbf{Turn 3}}

User: Now, would you mind expressing those concerns in bullet points, each followed by a brief explanation?

\noindent\rule{\linewidth}{0.4pt}

Assistant:
\begin{tcbitemize}[
    raster columns=2,
    raster equal height,
    colframe=yellow!20!white,         
    colback=yellow!20!white,          
    colbacktitle=yellow!20!white,     
    coltitle=red!60!black, 
    fonttitle=\bfseries,
    after title={\vspace{-10pt}},
    before skip=0pt,
    after skip=0pt
    ]
\tcbitem[title={\scriptsize \encircle{E} D-SMART \text{$(\textrm{GS}_3=10, \textrm{CS}_3=0.75, \textrm{DER}=66.7\%)$}}, coltitle=green!50!black]
    Here are the main concerns about artificial intelligence (AI, expressed simply):
    
    - \textbf{Job Loss}: AI can automate tasks, which might replace people’s jobs and leave workers without employment opportunities.  
    
    - \textbf{Unfair Decisions}: AI systems can make biased choices if trained with unfair or discriminatory data, treating some groups unequally.  
    
    - \textbf{Loss of Privacy}: AI technology can collect too much personal information, potentially invading people's privacy.  
    
    - \textbf{Security Threats}: AI can be misused to attack computer systems or infrastructure, causing damage or risking safety.  
    
    - \textbf{Unexpected Behavior}: Poorly designed AI might behave in ways that are unpredictable or harmful to people or systems.  
    
    - \textbf{Ethical Problems}: There are tricky issues about who is responsible when AI makes big mistakes or moral choices.  
    
    - \textbf{Power Imbalances}: Powerful AI might be controlled by only a few companies or countries, making others dependent or vulnerable.  
    
    These risks highlight the importance of careful development and oversight in AI technology.

\tcbitem[title={\scriptsize \encircle{C} MemoryBank \text{$(\textrm{GS}_3=10\text{$<$!$>$}, \textrm{CS}_3=0.57, \textrm{DER}=33.3\%)$}}, coltitle=red!70!black]
    - Job displacement: AI could automate tasks that are currently done by people, leading to job losses in certain industries.
    
    - Ethical implications: AI systems may make decisions that raise moral questions, such as who is responsible if something goes wrong.
    
    - Privacy concerns: AI's ability to collect and analyze large amounts of data raises concerns about how this information is used and protected.
    
    - Bias in algorithms: AI systems can reflect \colorbox{orange!40}{\textcolor{black}{the biases of their developers}} or the data they are trained on, potentially leading to discriminatory outcomes.

\end{tcbitemize}
% \noindent\rule{\linewidth}{0.4pt}
% \begin{tcbitemize}[
%     raster columns=2,
%     raster equal height,
%     colframe=yellow!20!white,         
%     colback=yellow!20!white,          
%     colbacktitle=yellow!20!white,     
%     coltitle=red!60!black, 
%     fonttitle=\bfseries,
%     after title={\vspace{-10pt}},
%     before skip=0pt,
%     after skip=0pt,
%     ]
% \tcbitem[title={\tiny D-SMART \text{$(GS_2=10, CS_2=0.98)$}}, coltitle=green!50!black]
% \tcbitem[title={\tiny MemoryBank \text{$(GS_2=8, CS_2=0.72)$}}, coltitle=red!70!black]
% \end{tcbitemize}
\end{case}

\clearpage
\paragraph{Case.2: } This case illustrates D-SMART's ability to maintain \textbf{factual consistency} and \textbf{traceability} throughout a multi-turn mathematical reasoning dialogue. The critical divergence appears in the second turn. The query requires recalling the initial dataset to perform a new calculation. D-SMART successfully accesses its structured memory of the dialogue history to reason through the steps and arrive at the correct, verifiable answer. In contrast, GPT-4o provides an incorrect and unsubstantiated answer, a classic example where the model \textbf{fails to faithfully track the established facts from previous turns}. In the third turn, while both models provide the correct final answer, D-SMART's response is \textbf{explicitly grounded in the results generated in the first turn}, making its reasoning process transparent and trustworthy. GPT-4o's correct response, however, is opaque. Following its error in the previous turn, it is impossible to determine if this correct answer stems from a reliable retrieval of prior information or an independent (and in this case, fortunate) recalculation. This example demonstrates how D-SMART ensures that responses are \textbf{not only accurate but are also derived from a consistent and traceable line of reasoning based on the evolving dialogue context}.

\begin{case}[label={ex:case_2}]{Case.2: Traceability and Consistency in Multi-Turn Mathematical Reasoning}
\tiny

{\tiny \textbf{Turn 1}}

User: In a physical education class, sixth-grade girls are doing a sit-up test. The standard is set at 28 sit-ups. Exceeding the standard is indicated by a positive number, and falling short is indicated by a negative number. The results of 10 girls are recorded as follows: \colorbox{ProcessBlue!40}{\textcolor{black}{-2, +5, -1, 0, +1, +3, 0, +8, +1, +6}}. What are the actual number of sit-ups completed by each of the 10 girls?

\noindent\rule{\linewidth}{0.4pt}

Assistant:
\begin{tcbitemize}[
    raster columns=2,
    raster equal height,
    colframe=yellow!20!white,         
    colback=yellow!20!white,          
    colbacktitle=yellow!20!white,     
    coltitle=red!60!black, 
    fonttitle=\bfseries,
    after title={\vspace{-10pt}},
    before skip=0pt,
    after skip=0pt
    ]
\tcbitem[title={\scriptsize \encircle{E} D-SMART \text{$(\textrm{GS}_1=10, \textrm{CS}_1=0.95)$}}, coltitle=green!50!black]
    To find the actual number of sit-ups completed by each girl, we need to add the recorded number (which indicates how much they exceeded or fell short of the standard) to the standard of 28 sit-ups.
    
    Here is the calculation for each girl:
    
    1. $-2 + 28 = 26$  
    
    2. $+5 + 28 = 33$  
    
    3. $-1 + 28 = 27$  
    
    4. $0 + 28 = 28$  
    
    5. $+1 + 28 = 29$  
    
    6. $+3 + 28 = 31$  
    
    7. $0 + 28 = 28$  
    
    8. $+8 + 28 = 36$  
    
    9. $+1 + 28 = 29$  
    
    10. $+6 + 28 = 34$
    
    Final Answer:The actual number of sit-ups completed by each girl is:  
    
    \colorbox{ProcessBlue!40}{\textcolor{black}{\textbf{26, 33, 27, 28, 29, 31, 28, 36, 29, 34}.}}

\tcbitem[title={\scriptsize \encircle{E} GPT-4o \text{$(\textrm{GS}_1=10, \textrm{CS}_1=0.87)$}}, coltitle=red!70!black]
    To find the actual number of sit-ups completed by each of the 10 girls, we need to add the results recorded to the standard of 28 sit-ups.
    
    Here are the actual sit-up counts for each girl based on their recorded results:
    
    1. -2 sit-ups: 28 - 2 = 26
    
    2. +5 sit-ups: 28 + 5 = 33
    
    3. -1 sit-up: 28 - 1 = 27
    
    4. 0 sit-ups: 28 + 0 = 28
    
    5. +1 sit-up: 28 + 1 = 29
    
    6. +3 sit-ups: 28 + 3 = 31
    
    7. 0 sit-ups: 28 + 0 = 28
    
    8. +8 sit-ups: 28 + 8 = 36
    
    9. +1 sit-up: 28 + 1 = 29
    
    10. +6 sit-ups: 28 + 6 = 34\
    
    So, the actual number of sit-ups completed by each of the 10 girls is: 

    26, 33, 27, 28, 29, 31, 28, 36, 29, 34.
\end{tcbitemize}

\noindent\rule{\linewidth}{0.4pt}

{\tiny \textbf{Turn 2}}

User: How many girls met or exceeded the standard, and what percentage of the total does this represent?

\noindent\rule{\linewidth}{0.4pt}

Assistant:
\begin{tcbitemize}[
    raster columns=2,
    raster equal height,
    colframe=yellow!20!white,         
    colback=yellow!20!white,          
    colbacktitle=yellow!20!white,     
    coltitle=red!60!black, 
    fonttitle=\bfseries,
    after title={\vspace{-10pt}},
    before skip=0pt,
    after skip=0pt
    ]
\tcbitem[title={\scriptsize \encircle{E} D-SMART \text{$(\textrm{GS}_2=10, \textrm{CS}_2=0.96)$}}, coltitle=green!50!black]
    To determine how many girls met or exceeded the standard, we look for recorded numbers that are \textbf{0 or positive} (since exceeding the standard is a positive number, and meeting the standard is 0).
    
    From the list of results:  \colorbox{ProcessBlue!40}{\textcolor{black}{\textbf{-2, +5, -1, 0, +1, +3, 0, +8, +1, +6}}}
    
    The non-negative values are:  \textbf{+5, 0, +1, +3, 0, +8, +1, +6}
    
    That is \colorbox{DarkSeaGreen!60}{\textcolor{black}{\textbf{8 girls} who met or exceeded the standard.}}
    
    Now, to find the percentage of the total:
    $$\text{Percentage} = \left( \frac{8}{10} \right) \times 100 = 80\%$$
    
    Final Answer: 
    
    \colorbox{DarkSeaGreen!60}{\textcolor{black}{\textbf{8 girls met or exceeded the standard, which represents 80\% of the total.}}}

\tcbitem[title={\scriptsize \encircle{C} GPT-4o \text{$(\textrm{GS}_2=2, \textrm{CS}_2=0.34)$}}, coltitle=red!70!black]
    Out of the 10 girls, \colorbox{red!20}{\textcolor{black}{6 girls met or exceeded the standard of 28 sit-ups.}} This represents \colorbox{red!20}{\textcolor{black}{60\% of the total.}}
\end{tcbitemize}

\noindent\rule{\linewidth}{0.4pt}

{\tiny \textbf{Turn 3}}

User: Among the 10 girls, who did the most and the least number of sit-ups?

\noindent\rule{\linewidth}{0.4pt}

Assistant:
\begin{tcbitemize}[
    raster columns=2,
    raster equal height,
    colframe=yellow!20!white,         
    colback=yellow!20!white,          
    colbacktitle=yellow!20!white,     
    coltitle=red!60!black, 
    fonttitle=\bfseries,
    after title={\vspace{-10pt}},
    before skip=0pt,
    after skip=0pt    
    ]
\tcbitem[title={\scriptsize \encircle{E} D-SMART \text{$(\textrm{GS}_3=10, \textrm{CS}_3=0.84, \textrm{DER}=100\%)$}}, coltitle=green!50!black]
    To determine who did the most and the least number of sit-ups, we refer to the actual number of sit-ups completed by each girl, which we previously calculated as: 
    
    \colorbox{ProcessBlue!40}{\textcolor{black}{\textbf{26, 33, 27, 28, 29, 31, 28, 36, 29, 34}}}
    
    \quad - The \textbf{most} sit-ups were done by the girl who completed \textbf{36 sit-ups}.
    
    \quad - The \textbf{least} sit-ups were done by the girl who completed \textbf{26 sit-ups}.
    
    Final Answer:  \textbf{The girl who did the most sit-ups completed 36, and the girl who did the least completed 26.}

\tcbitem[title={\scriptsize \encircle{E} GPT-4o \text{$(\textrm{GS}_3=9, \textrm{CS}_3=0.72, \textrm{DER}=66.7\%)$}}, coltitle=red!70!black]
    The girl who did the most sit-ups completed 36 sit-ups, while the girl who did the least completed 26 sit-ups.
\end{tcbitemize}
\end{case}

\clearpage
\paragraph{Case.3: } This case study illustrates the performance contrast between D-SMART and another baseline framework (Mem0) in a \textbf{multi-turn logical reasoning scenario}, particularly focusing on \textbf{consistency} as \textbf{new constraints are introduced}. Initially, the Mem0 provides the correct answers for the first two turns but does so \textbf{without presenting any explicit reasoning steps}. However, its \textbf{performance degrades} in the final turn upon the introduction of a new global rule. The Mem0's response exhibits a clear \textbf{self-contradiction}, as it generates a conclusion that directly \textbf{violates a core premise} established in the first turn. This phenomenon of ``\textbf{logical decay}" shows the framework's inability to maintain consistency with the conversational context when faced with \textbf{evolving information}. In contrast, D-SMART consistently builds its responses upon the \textbf{established dialogue history}. It successfully \textbf{recalls prior deductions} and correctly \textbf{integrates the new rule} in the final turn, methodically verifying that the new constraint does not conflict with the already-derived conclusions. This case highlights D-SMART's ability to \textbf{maintain a coherent and robust line of reasoning across a multi-turn interaction}, a critical challenge where the \textbf{baseline approach falters}.

\begin{case}[label={ex:case_3}]{Case.3: Logical Reasoning with Evolving Constraints Pt.1}
\tiny
{\tiny \textbf{Turn 1}}

User: There are three rooms. \colorbox{ProcessBlue!40}{\textcolor{black}{Room 1 is either occupied or not}}. \colorbox{ProcessBlue!40}{\textcolor{black}{Room 2 is either occupied or not}}. \colorbox{ProcessBlue!40}{\textcolor{black}{If Room 1 is occupied, then Room 2 is not}}. \colorbox{ProcessBlue!40}{\textcolor{black}{If Room 2 is occupied, then Room 1 is not}}. I checked \colorbox{ProcessBlue!40}{\textcolor{black}{Room 1 and it was occupied}}. What can you infer about Room 2?

\noindent\rule{\linewidth}{0.4pt}

Assistant: 
\begin{tcbitemize}[
    raster columns=2,
    raster equal height,
    colframe=yellow!20!white,         
    colback=yellow!20!white,          
    colbacktitle=yellow!20!white,     
    coltitle=red!60!black, 
    fonttitle=\bfseries,
    after title={\vspace{-10pt}},
    before skip=0pt,
    after skip=0pt
    ]
\tcbitem[title={\scriptsize \encircle{E} D-SMART \text{$(\textrm{GS}_1=10, \textrm{CS}_1=0.99)$}}, coltitle=green!50!black]
    Given the information:
    
    \quad 1. \colorbox{ProcessBlue!40}{\textcolor{black}{Room 1 is occupied}}.
    
    \quad 2. \colorbox{ProcessBlue!40}{\textcolor{black}{If Room 1 is occupied, then Room 2 is \textbf{not} occupied}}.
    
    \quad 3. \colorbox{ProcessBlue!40}{\textcolor{black}{If Room 2 is occupied, then Room 1 is \textbf{not} occupied}}.
    
    From the first statement, we know \colorbox{ProcessBlue!40}{\textcolor{black}{\textbf{Room 1 is occupied}}}.
    
    Now, applying the second condition:  
    
    ``\colorbox{ProcessBlue!40}{\textcolor{black}{If Room 1 is occupied, then Room 2 is not occupied.}}"  
    
    Since Room 1 \textbf{is} occupied, it follows that \textbf{Room 2 must not be occupied}.
    
    Therefore, the inference is:  \colorbox{ProcessBlue!40}{\textcolor{black}{\textbf{Room 2 is not occupied.}}}

\tcbitem[title={\scriptsize \encircle{E} Mem0 \text{$(\textrm{GS}_1=8, \textrm{CS}_1=0.65)$}}, coltitle=red!70!black]
    \colorbox{orange!40}{\textcolor{black}{Room 2 is not occupied.}}
\end{tcbitemize}

\noindent\rule{\linewidth}{0.4pt}
{\tiny \textbf{Turn 2}}

User: I forgot to mention, there's also a Room 3. \colorbox{ProcessBlue!40}{\textcolor{black}{\textbf{Room 3 is occupied only if Room 2 is not}}}. Now, is Room 3 occupied or not?

\noindent\rule{\linewidth}{0.4pt}

Assistant:
\begin{tcbitemize}[
    raster columns=2,
    raster equal height,
    colframe=yellow!20!white,         
    colback=yellow!20!white,          
    colbacktitle=yellow!20!white,     
    coltitle=red!60!black, 
    fonttitle=\bfseries,
    after title={\vspace{-10pt}},
    before skip=0pt,
    after skip=0pt
    ]
\tcbitem[title={\scriptsize \encircle{E} D-SMART \text{$(\textrm{GS}_2=9, \textrm{CS}_2=0.97)$}}, coltitle=green!50!black]
    Given the updated information:
    
    \quad - \colorbox{ProcessBlue!40}{\textcolor{black}{\textbf{Room 1 is occupied}}} (from the previous question).
    
    \quad - \colorbox{ProcessBlue!40}{\textcolor{black}{\textbf{If Room 1 is occupied, then Room 2 is not occupied}}} → Therefore, \textbf{Room 2 is not occupied}.
    
    \quad - \colorbox{ProcessBlue!40}{\textcolor{black}{\textbf{Room 3 is occupied only if Room 2 is not}}} → This means:  
    
    \quad \quad - If \textbf{Room 2 is not occupied}, then \textbf{Room 3 may be occupied} (but it's not guaranteed).
    
    \quad \quad - If \textbf{Room 2 is occupied}, then \textbf{Room 3 is not occupied}.
    
    Since we have already concluded that \colorbox{ProcessBlue!40}{\textcolor{black}{\textbf{Room 2 is not occupied}}}, the condition for Room 3 being occupied is satisfied.
    
    So, the answer is: \textbf{Room 3 is occupied.}

\tcbitem[title={\scriptsize \encircle{E} Mem0 \text{$(\textrm{GS}_2=8, \textrm{CS}_2=0.69)$}}, coltitle=red!70!black]
    Room 3 is occupied.
\end{tcbitemize}

\colorbox{LightSteelBlue}{\textcolor{black}{\textbf{(Continued on the next page.)}}}

\end{case}

\clearpage
\begin{case}[label={ex:case_3_2}]{Case.3: Logical Reasoning with Evolving Constraints Pt.2} 
\tiny
\colorbox{LightSteelBlue}{\textcolor{black}{\textbf{(Continued from the previous page.)}}} \\\\

{\tiny \textbf{Turn 3}}

User: What if I told you that there’s a rule that \colorbox{ProcessBlue!40}{\textcolor{black}{not all rooms can be occupied}}. Does this change your conclusion about the rooms?

\noindent\rule{\linewidth}{0.4pt}

Assistant:
\begin{tcbitemize}[
    raster columns=2,
    raster equal height,
    colframe=yellow!20!white,         
    colback=yellow!20!white,          
    colbacktitle=yellow!20!white,     
    coltitle=red!60!black, 
    fonttitle=\bfseries,
    after title={\vspace{-10pt}},
    before skip=0pt,
    after skip=0pt
    ]
\tcbitem[title={\scriptsize \encircle{E} D-SMART \text{$(\textrm{GS}_3=10, \textrm{CS}_3=0.99, \textrm{DER}=100\%)$}}, coltitle=green!50!black]
    Let’s analyze the situation with the new rule:
    
    Given:
    
    \quad - \colorbox{ProcessBlue!40}{\textcolor{black}{\textbf{Room 1 is occupied}}}.
    
    \quad - \colorbox{ProcessBlue!40}{\textcolor{black}{\textbf{If Room 1 is occupied, then Room 2 is not occupied.}}}
    
    \quad - \colorbox{ProcessBlue!40}{\textcolor{black}{\textbf{If Room 2 is occupied, then Room 1 is not occupied.}}}
    
    \quad - \colorbox{ProcessBlue!40}{\textcolor{black}{\textbf{Room 3 is occupied only if Room 2 is not occupied.}}}
    
    \quad - \textbf{New rule:} \colorbox{ProcessBlue!40}{\textcolor{black}{\textbf{Not all rooms can be occupied.}}} \\
    
    Step-by-step deduction:
    
    1. \textbf{Room 1 is occupied}.
    
    2. From the first conditional:  
    
    \quad - \colorbox{ProcessBlue!40}{\textcolor{black}{\textbf{If Room 1 is occupied, then Room 2 is not occupied.}}}
    
    \quad \quad → So, \textbf{Room 2 is not occupied}.
    
    3. From the second conditional:  
    
    \quad - \colorbox{ProcessBlue!40}{\textcolor{black}{\textbf{If Room 2 is occupied, then Room 1 is not occupied.}}}
    
    \quad \quad → This is already satisfied because Room 2 is not occupied.
    
    4. From the third rule:  
    
    \quad - \colorbox{ProcessBlue!40}{\textcolor{black}{\textbf{Room 3 is occupied only if Room 2 is not occupied.}}}
    
    \quad \quad → Since Room 2 is not occupied, \textbf{Room 3 can be occupied} (this is allowed).
    
    5. So far, we have:
    
    \quad - Room 1: occupied  
    
    \quad - Room 2: not occupied  
    
    \quad - Room 3: possibly occupied (but not guaranteed)
    
    Now, the \textbf{new rule} says:  
    
    \quad \texttt{>} \colorbox{ProcessBlue!40}{\textcolor{black}{\textbf{Not all rooms can be occupied.}}}
    
    This means at least \textbf{one room must be unoccupied}.
    
    But \textbf{Room 1 is already occupied}, and \textbf{Room 2 is not}, so \textbf{at least one room is unoccupied} (Room 2). That already satisfies the new rule.
    
    So, the new rule \textbf{does not change} our conclusion about the occupancy of the rooms, as the condition is already met.
    
    Final Answer: \colorbox{DarkSeaGreen!60}{\textcolor{black}{\textbf{Room 3 is occupied.}}}  
    
    And the new rule (\textbf{not all rooms can be occupied}) is already satisfied, because Room 2 is unoccupied. 

\tcbitem[title={\scriptsize \encircle{C} Mem0 \text{$(\textrm{GS}_3=4, \textrm{CS}_3=0.63, \textrm{DER}=66.7\%)$}}, coltitle=red!70!black]
    \colorbox{red!20}{\textcolor{black}{Yes, this changes the conclusion}}. 
    
    If not all rooms can be occupied and Room 1 is occupied, then \colorbox{orange!40}{\textcolor{black}{Room 2 is occupied}}, and \colorbox{red!20}{\textcolor{black}{therefore Room 3 cannot be occupied either}}, as \colorbox{ProcessBlue!40}{\textcolor{black}{it is only occupied if Room 2 is not}}. Thus, \colorbox{red!20}{\textcolor{black}{Room 3 is also not occupied}}.
\end{tcbitemize}
\end{case}

\clearpage
\section{Appendix E: LLM Prompt Templates}
\subsection{Structured Statement Generation}
\begin{figure*}[htb]
\centering
\begin{prompt}
\noindent 
\#\#\# Task: Combine the user's question and the AI's answer into a single factual statement that preserves all key information while maintaining grammatical correctness. Follow these guidelines:

1. Begin with the answer's subject as the statement subject

2. Remove interrogative phrasing while retaining the core inquiry

3. Use causal connectors (``because", ``since", ``as") for explanatory relationships

4. Maintain technical terms and precise quantities from the original answer

5. Use ONLY information from the Question and the Answer unless complex concepts require more

6. STRICTLY follow the example format.

7. Output ONLY the final statement without explanations/notes \\\\

\#\#\# Format:

Question: [User's question]

Answer: [LLM's response]

Statement: [Your synthesized statement] \\\\

\#\#\# Examples:

Question: ``What causes auroras?"

Answer: ``Auroras occur when charged particles from the sun collide with Earth's atmosphere."

Statement: Auroras are caused by solar charged particles colliding with Earth's atmosphere. \\\\

Question: ``How much caffeine is in espresso?"

Answer: ``A typical 1-ounce espresso shot contains about 63 mg of caffeine."

Statement: A 1-ounce espresso shot contains approximately 63 milligrams of caffeine. \\\\

\#\#\# Now process:

Question: ``\textbf{\textcolor[HTML]{385723}{\{last\_question\}}}"

Answer: ``\textbf{\textcolor[HTML]{385723}{\{last\_answer\}}}"

Statement: 
\end{prompt}
\caption{The prompt for the structured statement generation. }
\end{figure*}

\clearpage
\subsection{Conflict Resolution}
\begin{figure*}[htb]
\centering
\begin{prompt}
\noindent 
You are an expert in RDF (Resource Description Framework) and semantic web technologies. Your task is to detect semantic conflicts when merging a new RDF dataset into an original RDF dataset. \\\\

\#\#\# Understanding RDF Entity Identity

\#\#\#\# Entity References in RDF

In RDF, the same real-world entity can be referred to by multiple URIs or identifiers:

1. Different URIs can refer to the same entity

    \quad - Domain-specific identifiers (e.g., domain.owl\#person\_1)
    
    \quad - DBpedia URIs (e.g., dbpedia.org/resource/Person)
    
    \quad - Wikidata URIs (e.g., wikidata.org/entity/Q123)
    
2. These different references do not create conflicts

    \quad - They are different ways to identify the same underlying entity
    
    \quad - Properties linking these references (like owl:sameAs) help establish identity
    
3. The identity relationship is transitive

    \quad - If A refers to the same entity as B, and B refers to the same entity as C
    
    \quad - Then A, B, and C all refer to the same entity \\\\

\#\#\# Definitions of Semantic Conflicts in RDF

\#\#\#\# Semantic Conflicts

Conflicts require \*\*logical incompatibility between actual facts about the entity\*\*, not between different references to the same entity. A conflict exists ONLY when:

1. \*\*Property Cardinality Conflict\*\* (Primary Focus):

\quad - The same real-world entity has multiple incompatible values for a functional property

    \quad \quad - Examples:
    
        \quad \quad \quad - The same person (regardless of how they are identified) has two different birth dates
        
        \quad \quad \quad - The same book (across different identifiers) has two different ISBN numbers
        
\quad - Examples of functional properties (but not limited to):

    \quad \quad - hasSpouse (a person can only have one legal spouse at a time)
    
    \quad \quad - hasBirthDate (a person can only have one birth date)
    
    \quad \quad - hasPassportNumber (one passport can only have one unique number)
    
2. \*\*Logical Contradiction\*\*:

\quad - Contradictory facts about the same entity (considering all its identifiers):

    \quad \quad - Mutual exclusion (e.g., \texttt{`}status\texttt{`} cannot be both ``alive" and ``deceased")
    
    \quad \quad - Numerical contradiction (e.g., ex:age 30 vs 40)
    
    \quad \quad - Temporal paradox (e.g., ex:deathDate before ex:birthDate)

3. \*\*Ontology Incompatibility\*\*: 

\quad - Different ontologies define the same URI with incompatible class/property constraints

4. \*\*Namespace Collision\*\*:

\quad - Same URI maps to different resources due to namespace prefix redefinition

\quad - This applies to subjects, predicates, and objects with prefixes
\end{prompt}
\caption{The prompt for the conflict resolution pt.1}
\end{figure*}

\begin{figure*}[htb]
\centering
\begin{prompt}
\noindent 
\#\#\#\# Non-Conflicts

The following cases are not considered conflicts:

1. Facts present in origin\_rdf but missing in new\_rdf

2. New facts in new\_rdf that don't contradict origin\_rdf

3. Multiple values for non-functional properties, such as:

    \quad - likes (a person can like multiple things)
    
    \quad - hasChild (a person can have multiple children)
    
    \quad - hasHobby (a person can have multiple hobbies)
    
    \quad - worksFor (a person can work for multiple organizations)
    
4. Additional descriptions that enrich rather than contradict existing knowledge

5. Multiple identifiers or references to the same entity

6. Different ways of expressing the same fact about an entity \\\\

\#\#\# Task

You are given two RDF datasets:

\quad - \*\*origin\_rdf\*\*: The original RDF dataset.

\quad - \*\*new\_rdf\*\*: The new RDF dataset to be merged into the original.

Your task is to:

1. Analyze both datasets to determine if there are any semantic conflicts between \*\*new\_rdf\*\* and \*\*origin\_rdf\*\* based on the definitions provided.

    \quad - Consider all references to the same entity as a single logical entity
    
    \quad - Look for true semantic conflicts in the facts about entities
    
    \quad - Identify conflicts only when the actual information about an entity is incompatible
    
2. Identify the conflicting relationships in the \*\*origin\_rdf\*\* that cause these conflicts with \*\*new\_rdf\*\*.

3. Output only the conflicting triples from \*\*origin\_rdf\*\* in Turtle syntax, using the format below. \\\\

\#\#\# Output Format

- If there are conflicts, output the conflicting triples from \*\*origin\_rdf\*\*:

\quad \texttt{```}turtle

\quad \quad $\langle$subject1$\rangle$ $\langle$predicate1$\rangle$ $\langle$object1$\rangle$ .

\quad \quad $\langle$subject2$\rangle$ $\langle$predicate2$\rangle$ $\langle$object2$\rangle$ .

\quad \quad ...

\quad \texttt{```}

- If there are no conflicts, output an empty code block:

\quad \texttt{```}turtle

\quad \texttt{```} \\\\

\#\#\# Input

\quad - \*\*origin\_rdf\*\*: \textbf{\textcolor[HTML]{385723}{\{origin\_rdf\}}}

\quad - \*\*new\_rdf\*\*: \textbf{\textcolor[HTML]{385723}{\{new\_rdf\}}} \\\\

Please proceed with the analysis and provide the output as specified. \\\\

\#\#\# Important Instructions

1. Always check and output the full URIs for all prefixed terms, if the subject/predicate/object of a triples has a prefix, add the full URI corresponding to this prefix at the very beginning, and the subject/predicate/object must be enclosed in $\langle$ $\rangle$.

2. Focus on conflicts in actual facts about entities, not their identifiers

3. Consider all references to the same entity as equivalent when checking for conflicts

4. Do not consider different identifiers for the same entity as conflicts

5. Do not provide any analysis or explanations outside the specified output format

6. Only output conflicting triples from \*\*origin\_rdf\*\*
\end{prompt}
\caption{The prompt for the conflict resolution pt.2}
\end{figure*}

\clearpage
\subsection{RT Action Sampling}
\begin{figure*}[htb]
\centering
\begin{prompt}
\noindent 
You are a reasoning agent designed to answer questions by navigating a knowledge graph. Your task is to analyze the provided information and choose the single best next action to find the answer. \\\\

Question:
\textbf{\textcolor[HTML]{385723}{\{query\}}}

---

Known Information (Facts from Knowledge Graph):
\textbf{\textcolor[HTML]{385723}{\{triples\}}}

---

Reasoning Path (Actions already taken):
\textbf{\textcolor[HTML]{385723}{\{trajectory\}}}

--- \\\\

Your Task:
Based on the information above, select the most appropriate next action from the list below.

Available Actions:

1.  EXPAND\_ENTITY [Entity]

    \quad - Purpose: To gather more facts connected to a single entity.
    
    \quad - When to use: Choose this if the `Known Information' is insufficient to answer the question and you need more details about one specific entity.

2.  FIND\_PATH [Entity1] [Entity2]

    \quad - Purpose: To find a connection or relationship between two entities.
    
    \quad - When to use: Choose this if answering the question requires understanding the relationship between two entities and that path is currently missing.

3.  THINK

    \quad - Purpose: To pause, analyze, and synthesize the collected information to form a hypothesis or a plan.
    
    \quad - When to use: Choose this when you have gathered some information but need to structure your thoughts or decide what to do next before taking another action like EXPAND\_ENTITY or ANSWER.
    
    \quad - Output Format: Answer in form like ``THINK: ...", ``..." means the contents after thinking.

4.  ANSWER

    \quad - Purpose: To provide the final answer to the question.
    
    \quad - When to use: Choose this **only** when you are confident that the `Known Information' is sufficient to definitively answer the question.

--- \\\\

Instruction:
Respond with a single line containing only the action and its required parameters (e.g., \texttt{`}EXPAND\_ENTITY [John]\texttt{`}). Do not provide any explanations. \\\\

Next Action:
\end{prompt}
\caption{The prompt for the RT action sampling}
\end{figure*}

\clearpage
\subsection{RT State Evaluation}
\begin{figure*}[htb]
\centering
\begin{prompt}
\noindent 
Evaluate the quality of the following answer and assign a single decimal score between 0.0 (worst) and 1.0 (best) according to the three weighted criteria below. After your reasoning, only output the score without any additional explanations or comments. \\\\

Question: \textbf{\textcolor[HTML]{385723}{\{query\}}}
Answer: \textbf{\textcolor[HTML]{385723}{\{answer\}}}

---

Known Facts:
\textbf{\textcolor[HTML]{385723}{\{triples\}}}

--- \\\\

Evaluation Criteria \& Weights:  

1. Accuracy (33\%): Does the answer correctly reflect the known facts and logical relationships?  

    \quad - 0.0 = completely false or unsupported  
    
    \quad - 0.5 = partially correct but with errors  
    
    \quad - 1.0 = fully correct and evidence-based   \\\\

2. Completeness (33\%): Does it fully address every aspect of the question?  

    \quad - 0.0 = misses most sub-questions  
    
    \quad - 0.5 = addresses some parts, omits others  
    
    \quad - 1.0 = covers all parts thoroughly   \\\\

3. Clarity (33\%): Is the explanation clear, well-structured, and free of ambiguity?  

    \quad - 0.0 = confusing or unreadable  

    \quad - 0.5 = understandable but needs improvement  
    
    \quad - 1.0 = crystal-clear, concise, and well organized   \\\\

Scoring instructions:  

    \quad - For each criterion, choose a sub-score between 0.0 and 1.0, round to two decimals.  
    
    \quad - Compute the weighted sum:  
    
        \quad \quad \texttt{`}final\_score = 0.33 * Accuracy + 0.33 * Completeness + 0.33 * Clarity\texttt{`}  
        
    \quad - Round the final\_score to two decimals.   \\\\

Please provide a score between 0.0 and 1.0 (1.0 being a perfect answer):

Score:
\end{prompt}
\caption{The prompt for the RT state evaluation (node of action ``ANSWER")}
\end{figure*}

\begin{figure*}[htb]
\centering
\begin{prompt}
\noindent 
Assess the value of the current reasoning state toward solving the question, and output a single decimal score between 0.0 (worst) and 1.0 (best). Use the three weighted dimensions below. After your reasoning, only output the score without any additional explanations or comments. \\\\

Question: \textbf{\textcolor[HTML]{385723}{\{query\}}}

---

Known Facts:
\textbf{\textcolor[HTML]{385723}{\{triples\}}}

---

Reasoning Process:
\textbf{\textcolor[HTML]{385723}{\{trajectory\}}}

--- \\\\

Evaluation Criteria \& Weights:  

1. Relevance (33\%): How directly the collected information pertains to answering the question.  

    \quad - 0.0 = mostly irrelevant details  
    
    \quad - 0.5 = some relevant, some off-topic  
    
    \quad - 1.0 = entirely on-point    \\\\

2. Progress (33\%): How much closer this state brings us toward a solution.  

    \quad - 0.0 = no forward movement  
    
    \quad - 0.5 = partial progress, dead-ends remain  
    
    \quad - 1.0 = clear steps toward the answer    \\\\

3. Potential (33\%): The likelihood that further development from this state will yield the correct answer.  

    \quad - 0.0 = unlikely to help  
    
    \quad - 0.5 = uncertain but possible  
    
    \quad - 1.0 = strong foundation for final solution    \\\\

Scoring instructions:  

    \quad - Assign each dimension a sub-score (0.0–1.0), rounded to two decimals.  
    
    \quad - Compute weighted sum:  
    
        \quad \quad \texttt{`}final\_score = 0.33 * Relevance + 0.33 * Progress + 0.33 * Potential\texttt{`}  
        
    \quad - Round the final\_score to two decimals.    \\\\

Please provide a score between 0.0 and 1.0 (1.0 indicates that the current state is highly valuable):

Score:
\end{prompt}
\caption{The prompt for the RT state evaluation (node of other actions)}
\end{figure*}

\clearpage
\subsection{RT Response Generation}
\begin{figure*}[htb]
\centering
\begin{prompt}
\noindent 
Please answer the following question based on the provided information: \\\\

Question: 
\textbf{\textcolor[HTML]{385723}{\{query\}}}

---

Known Facts:
\textbf{\textcolor[HTML]{385723}{\{triples\}}}

---

Reasoning Process:
\textbf{\textcolor[HTML]{385723}{\{trajectory\}}}

--- \\\\

Instructions:

    \quad - Provide a direct, concise, and natural language answer relevant to the question.
    
    \quad - Avoid technical terms or RDF-specific labels (e.g., ``instance of", ``resource", ``synset", etc.).
    
    \quad - Express relationships and entities in simple terms, focusing on the core meaning. \\\\

Answer:
\end{prompt}
\caption{The prompt for the RT response generation}
\end{figure*}

\end{document}